\documentclass[pageno]{jpaper}

\usepackage[normalem]{ulem}
\usepackage{xspace}
\usepackage{subfig}
\usepackage{lipsum}
\usepackage{varwidth}
\usepackage{array}
\usepackage{url}
\usepackage{xcolor}
\usepackage{amssymb}
\usepackage{pifont}
\usepackage{flushend}
\usepackage{authblk}

\usepackage{tikz}

\newcommand{\cmark}{\textcolor{green}{\ding{52}}}%
\newcommand{\xmark}{\textcolor{red}{\ding{55}}}%

\newcommand{\autop}[0]{AutoPilot\xspace}
\newcommand{\Fig}[1]{Fig.~\ref{#1}}
\newcommand{\red}[1]{\textcolor{red}{#1}}
\newcommand*\numcircledmod[1]{\raisebox{.5pt}{\textcircled{\raisebox{-.9pt} {#1}}}}

\begin{document}

\title{AutoPilot: Automating Co-Design Space Exploration for Autonomous UAVs}

\author[$\dagger$]{Srivatsan~Krishnan}
\author[$\dagger$]{Zishen~Wan}
\author[$\dagger$]{Kshitij~Bhardwaj}
\author[$\mp$]{Paul~Whatmough}
\author[$\S$]{Aleksandra Faust}
\author[$\dagger$]{Sabrina M. Neuman} 

\author[$\dagger$]{ Gu-Yeon~Wei}
\author[$\dagger$]{David~Brooks}
\author[$\dagger$]{Vijay Janapa Reddi}

\affil[$\dagger$]{Harvard University}
\affil[$\mp$]{ARM Research}
\affil[$\S$]{Google Brain Research}

\date{}
\maketitle

\begin{abstract}
\vspace{10pt}
Building domain-specific accelerators for autonomous unmanned aerial vehicles (UAVs) is challenging due to a lack of systematic methodology for designing onboard compute. Balancing a computing system for a UAV requires considering both the cyber (e.g., sensor rate, compute performance) and physical (e.g., payload weight) characteristics that affect overall performance. Iterating over the many component choices results in a combinatorial explosion of the number of possible combinations: from 10s of thousands to billions, depending on implementation details. Manually selecting combinations of these components is tedious and expensive. To navigate the {cyber-physical design space} efficiently, we introduce \emph{AutoPilot}, a framework that automates full-system UAV co-design. AutoPilot uses Bayesian optimization to navigate a large design space and automatically select a combination of autonomy algorithm and hardware accelerator while considering the cross-product effect of other cyber and physical UAV components. We show that the \autop methodology consistently outperforms general-purpose hardware selections like Xavier NX and Jetson TX2, as well as dedicated hardware accelerators built for autonomous UAVs, across a range of representative scenarios (three different UAV types and three deployment environments). Designs generated by \autop increase the number of missions on average by up to  2.25$\times$, 1.62$\times$, and 1.43$\times$ for nano, micro, and mini-UAVs respectively over baselines.
Our work demonstrates the need for holistic full-UAV co-design to achieve maximum overall UAV performance and the need for automated flows to simplify the design process for autonomous cyber-physical systems.
\end{abstract}

\section{Introduction}
\label{sec:intro}

Unmanned aerial vehicles (UAVs) are on the rise in real-world deployments~\cite{Timothy2017,medical-org,8373043,search-and-rescue-org}, but building computing systems for these platforms remains challenging. They are complex systems in which the traditional computing platform is just one component among many others. To achieve overall performance, it is important to understand what implications other UAV components have on the design of onboard compute.




\begin{figure}
\vspace{5pt}
        \includegraphics[width=0.99\columnwidth]{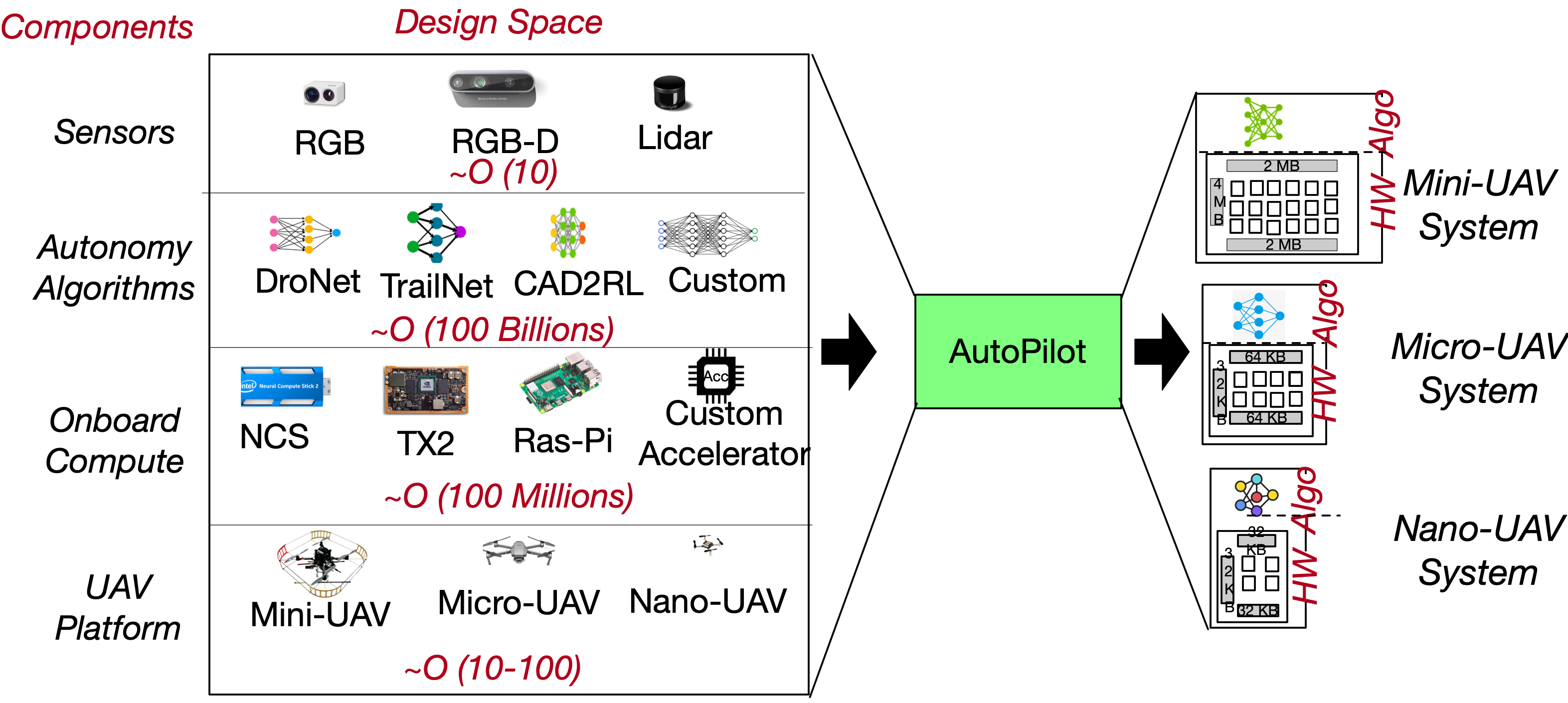}
        \caption{AutoPilot efficiently automates navigation of the large UAV component design space to co-design optimal onboard compute across a range of autonomous UAV systems.\vspace{-2pt}}
        \label{fig:uav-full-stack}
\end{figure}

Co-designing hardware accelerators with other UAV components requires navigating a large design space (see \Fig{fig:uav-full-stack}), e.g., \textit{100's} of UAVs~\cite{uav-count} $\times$ \textit{millions} of HW accelerators~\cite{accelerator-count} $\times$ \textit{billions} of autonomy algorithm neural network model paramaters~\cite{elsken2019neural} $\times$ \textit{100's} of sensors~\cite{sensor-count} $\approx$ 10\textsuperscript{18}. Worse, this number is still conservative since each UAV type includes additional components such as a flight controller and a battery. Taming this large space can be expensive and tedious. Automating the co-design of the hardware accelerator and other UAV system components can optimize mission performance while keeping design overheads low as UAV systems evolve.

Key challenges in UAV design include
the ability to systematically navigate the large design space of components, and understanding which combinations of these components maximize overall UAV performance. While specialized hardware is critical for compute efficiency, designing it is an expensive process. It is essential to establish automated design methodologies that remain agile as future autonomous systems evolve.

To address these challenges, we introduce
\emph{AutoPilot: a cyber-physical co-design automation framework for autonomous UAVs}. Given a high-level specification of autonomy task, UAV type, and mission goals, \autop automatically navigates the large design space to perform full-system UAV co-design to generate a combination of autonomy algorithm and corresponding hardware accelerator to maximize overall UAV performance (e.g., number of missions).

\begin{figure}[t!]
        \vspace{5pt}
        \centering
        \includegraphics[width=0.815\columnwidth]{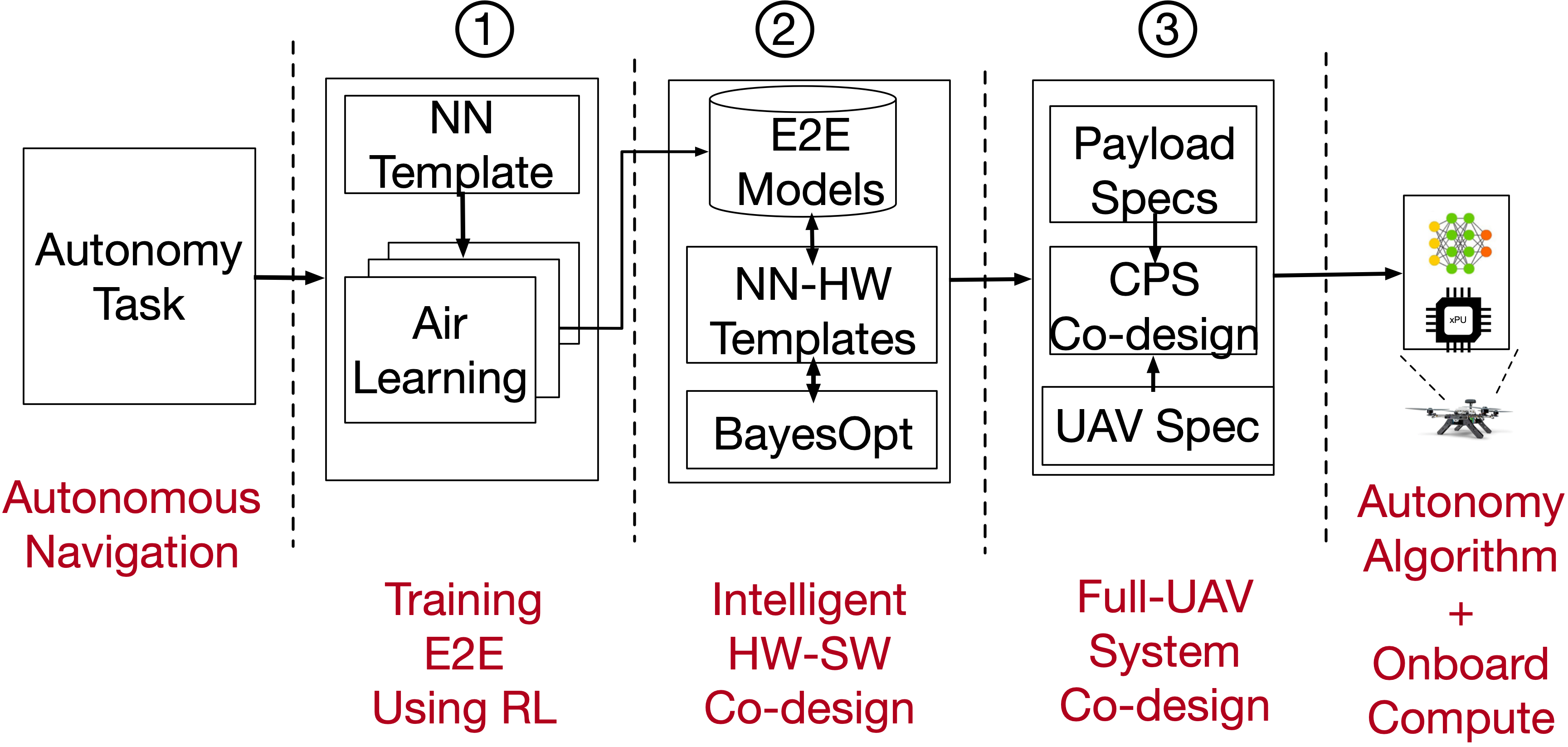}
        \caption{Overview of \autop design flow.\vspace{-10pt}}
        \label{fig:autopilot-high-level}
\end{figure}

The \autop takes
a high-level specification for the autonomy task as input, and has three main steps (see \Fig{fig:autopilot-high-level}):
(1) Trains several end-to-end autonomy algorithms for a given autonomy task (autonomous navigation) using reinforcement learning~\cite{airlearning}, and validate task-level functionality  (\numcircledmod{1} in \Fig{fig:autopilot-high-level}).
(2) Perform multi-objective algorithm-hardware co-design to maximize task success rate, compute performance and minimize power using Bayesian optimization~\cite{bayesopt}  (\numcircledmod{2} in \Fig{fig:autopilot-high-level}). This ensures that we traverse the large design space rapidly and obtain several interesting design candidates.
(3) Finally, we perform co-design across the full system of UAV components to select the optimal compute and autonomy algorithm combination for a given UAV to maximize its mission performance, i.e., number of missions  (\numcircledmod{3} in \Fig{fig:autopilot-high-level}). This step accounts for the cross-product effect across the full-UAV stack (\Fig{fig:uav-full-stack}) and is critical to maximize UAV performance.

Prior work has largely focused on quantifying the UAV design space~\cite{aspolos-drone} and designing standalone UAV compute systems~\cite{robox,pulp-dronet} (see Table~\ref{tab:related-work}).
Work examining the entire cyber-physical UAV system stack has quantified tradeoffs~\cite{aspolos-drone} but does not provide a systematic automated solution to navigate this enormous design space to produce optimal designs. 
Previous work also focuses on optimizing single kernels (e.g., localization and mapping~\cite{aspolos-drone}),
while a fully autonomous UAV requires the synthesis of many computationally-intensive kernels (e.g., map integration, motion planning)~\cite{mavbench,high-speed-drone,high-speed-ppc}.
Prior work on designing UAV compute has provided solutions using off-the-shelf hardware~\cite{source-seeking,high-speed-drone,gao2021ielas,wan2021survey,liu2021robotic,dronet,cad2rl,trailnet} or standalone hardware designed in isolation from the rest of the cyber-physical stack~\cite{pulp-dronet,navion}.

By contrast, with AutoPilot, we offer a complete solution for \emph{automatically navigating} the UAV design space and performing \emph{holistic co-design} across the entire UAV system stack.
To demonstrate the scalability of the \autop methodology, we apply it to three different UAV types and three domain randomized environments (total of nine combinations).
These auto-generated domain randomized environments have varying degrees of obstacle densities. They represent common deployment use cases for drones. For instance, autonomous navigation for a farming use case could be very sparse (low-obstacle), whereas a search and rescue operation in a forest is a dense-obstacle scenario.
The three fully randomized environments represent the deployment scenario ranging from low obstacle density to high obstacle density (to signify the difficulty in the autonomous task).  
We show that the \autop methodology consistently outperforms general-purpose hardware selections like Xavier NX and Jetson TX2 and dedicated hardware accelerators built for autonomous UAVs across a range of representative scenarios.

In summary, our main contributions are as follows:
\begin{itemize}
    \item Introduce \autop, an intelligent \emph{design space exploration framework} that \emph{automatically} co-designs a learning-based autonomy algorithm and its accelerator for a diverse class of UAVs and deployment scenarios.
    \item Leverage \emph{cyber-physical co-design} for UAVs, taking the entire system stack under consideration to maximize overall UAV performance, i.e., number of missions.
    \item Generate optimal designs that increase the number of missions by 2.25$\times$, 1.62$\times$, and 1.43$\times$ for nano, micro, and mini-UAVs respectively over baselines.
\end{itemize}

\begin{table}[t!]
\centering
\renewcommand\arraystretch{1.6}
\resizebox{1.0\columnwidth}{!}{%
\Huge
\begin{tabular}{|c|c|c|c|c|c|c|}
\hline
\multirow{2}{*}{\textbf{\begin{tabular}[c]{@{}c@{}}Prior \\ Work\end{tabular}}} & \multirow{2}{*}{\textbf{\begin{tabular}[c]{@{}c@{}}End-to-End\\ Autonomy?\end{tabular}}} & \multirow{2}{*}{\textbf{\begin{tabular}[c]{@{}c@{}}Hardware\\ Acceleration\end{tabular}}} & \multicolumn{2}{c|}{\textbf{Cyber and Physical Parameters}} & \multirow{2}{*}{\textbf{\begin{tabular}[c]{@{}c@{}}Provides\\ Design\\ Methodology?\end{tabular}}} & \multirow{2}{*}{\textbf{Automated?}} \\ \cline{4-5}
 &  &  & \textbf{\begin{tabular}[c]{@{}c@{}}Varies Sensor\\ (Frame rates)\end{tabular}} & \textbf{\begin{tabular}[c]{@{}c@{}}Varies Physical\\ parameters \\ (Thrust-to-ratio)\end{tabular}} &  &  \\ \hline
Navion & \xmark & Only VIO & \xmark & \xmark & \xmark & \xmark \\ \hline
Hadidi et al & \xmark & Only SLAM & \xmark & \xmark & \cmark & \xmark \\ \hline
RoboX & \xmark & Only Motion Planning & \xmark & \cmark & \cmark & \cmark \\ \hline
MavBench & \cmark & \xmark & \xmark & \xmark & \xmark & \xmark \\ \hline
PULP-DroNet & \cmark & Full end-to-end stack & \xmark & \xmark & \xmark & \xmark \\ \hline

\noalign{\hrule height 6pt}
\multicolumn{1}{!{\vrule width 6pt}c|}{
\begin{tabular}[c]{@{}c@{}c@{}}AutoPilot\\ (This Work)\end{tabular}}& \cmark & Full End-to-End Stack & 
\begin{tabular}[c]{@{}c@{}}\cmark\\ (Section~\ref{sec:sensor-performance})\end{tabular} & 
\begin{tabular}[c]{@{}c@{}}\cmark\\ (Section~\ref{sec:uav-physics})\end{tabular} &
\begin{tabular}[c]{@{}c@{}}\cmark\\ (Section~\ref{sec:uav-physics})\end{tabular} &
\multicolumn{1}{|c!{\vrule width 6pt}}{
\begin{tabular}[c]{@{}c@{}}\cmark \end{tabular}}\\ \hline
\noalign{\hrule height 6pt}
\end{tabular}}
\caption{Comparison of selected related work. \autop provides an automated methodology for co-design across the cyber-physical system stack for end-to-end UAV autonomy.\vspace{-10pt} }
\label{tab:related-work}
\end{table}

\section{Background}
\label{sec:background}

Autonomous UAVs have many components that must be considered in the co-design process (\Fig{fig:uav-full-stack}). We first define UAV size classifications. We then describe the basic components of a canonical UAV (without full autonomy), and finally, we discuss additional components needed for achieving autonomy.

\noindent \textbf{UAV Size Classifications.} UAV size impacts its physics, which must be considered for full cyber-physical co-design. There are three broad categories of UAV sizes: mini-UAVs (\textit{100's-1000's mm \& 1-1.5 Kgs}), micro-UAVs (\textit{100's mm \& 100's g}), and nano-UAVs (\textit{10's mm \& 10's g}).
UAV size determines the maximum payload weight and battery capacity.

\subsection{Base UAV Components}
A base UAV system contains components that allow humans to teleoperate the UAV but does not have the necessary components for full autonomy.

\textbf{Electro-Mechanical Components.} Depending upon the UAV size, the UAV has a proportional size frame where all the other components are mounted. The propulsion uses motors that share the battery energy and other electronics in the UAV. Electronic speed control (ESC) is used to provide precise rotor speed based on the commands from the flight controller.

\textbf{Flight Controller.} Flight controller is solely dedicated to the stabilization and control of the UAV. The flight controller firmware is computationally lightweight and is typically run on microcontrollers~\cite{ucontroller-fc-1,ucontroller-fc-2}, that are tightly integrated into the UAV platform. The flight controller uses onboard sensors, such as the Inertial Measurement Unit (IMU)~\cite{imu} and GPS, to stabilize and control the UAV. To recover from unpredictable errors (sudden winds or damaged rotors), the inner-loop typically runs at closed-loop frequencies of up to 1 kHz~\cite{1khz-control,koch2019neuroflight}.



\textbf{Battery.} The battery capacity of a UAV is the limiting factor in the number of missions it can complete on a single charge before its energy budget is depleted. UAV size and payload restrictions determine its battery capacity. Hence, flight time is often limited to few tens of minutes. Typically, a mini-UAVs can carry a bigger battery (1000's of mAh for about 20 to 25 minutes of flight) due to their larger size. 
Nano-UAVs, on the other hand, due to their small size, can only carry very small batteries (100's mAh for about 6 to 7 minutes of flight time).  

\subsection{Additional Components for UAV Autonomy}
\label{sec:algorithms}
To achieve full autonomy, the base UAV system needs to be augmented with more components to allow the UAV to make intelligent decisions without any human intervention.


\textbf{External Sensors.} The sensor creates snapshots of the environment and feeds the information to the autonomy algorithm (see below), generating high-level action. The UAV size and weight restriction impact the selection of sensors. The most common type of sensors (across UAV types) is RGB camera~\cite{lange2012autonomous,marcial2019estimation}. The weight of the sensor (payload weight), its performance (framerate), and sensing range affect the UAV's ability to navigate an environment autonomously.



\textbf{Autonomy Algorithm.} Autonomy in drones can be achieved by algorithms classified into two broad categories: Sense-Plan-Act (SPA) and End-to-End (E2E) Learning. 

In \textit{SPA}  the algorithm is broken into distinct stages: sensing, planning, and control.
In the sensing stage, sensor data is used to create a map~\cite{rusu20113d,elfes1989using,dissanayake2001solution} of the environment. The planning stage~\cite{rrt,motion-planning-survey} processes the map to determine the best trajectory. The trajectory information is used by the control stage, which actuates the rotor, so the robot follows the trajectory. SPA is usually slow in performance~\cite{dronet,pulp-dronet, source-seeking,sniffy-bug}. 

\textit{E2E} learning is an alternate paradigm where the algorithms process raw input sensor information (e.g., RGB, Lidar) and use a neural network model to directly produce output actions.
Unlike SPA, the E2E learning methods do not require maps or separate planning stages and hence are much faster than non-NN-based autonomy algorithms (i.e., SPA)~\cite{pulp-dronet,vgg-tx2}. The model can be trained using supervised learning~\cite{e2e-nvidia,dronet,trailnet0,trail-net,tesla-dnn} or reinforcement learning~\cite{source-seeking,cad2rl, prm-rl,qt-opt,rl-car,quarl,anwar2020autonomous}. Major companies working on autonomy have made public announcements about using E2E-based approaches~\cite{tesla-dnn,mobileye-drl}.

\textbf{Onboard Compute.} Autonomous UAVs have a dedicated compute subsystem that runs the autonomy algorithm, generating high-level actions or plans. Onboard compute can be classified into two types: off-the-shelf general-purpose compute and domain-specific hardware accelerators.
Due to their widespread availability, prior work focusing on autonomy algorithms often rely on general-purpose hardware such as Intel NUC, Nvidia Jetson TX1/2, ARM Cortex 9, or even microcontrollers like M4~\cite{crazyflie} to validate the functionality of the autonomy algorithms. The selection criteria are often based on the size/weight of the UAVs. Due to their larger payload carrying capability, mini-UAVs typically use general-purpose compute like Intel NUC, Jetson TX1/TX2. Nano-UAVs, due to their limited size, have microcontrollers~\cite{crazyflie}.
Recently there has been efforts to build domain-specific accelerators for autonomous UAVs (see Section~\ref{sec:related} for details).
However, these accelerators are typically optimized for higher throughput or low power without consideration of how these performance gains affect the compute weight. The payload weight of these specialized accelerators is a critical metric to consider for UAV system co-design because it affects a UAV's agility~\cite{f-1}.



\begin{figure*}[h!]
\centering
  \includegraphics[width=\textwidth,keepaspectratio]{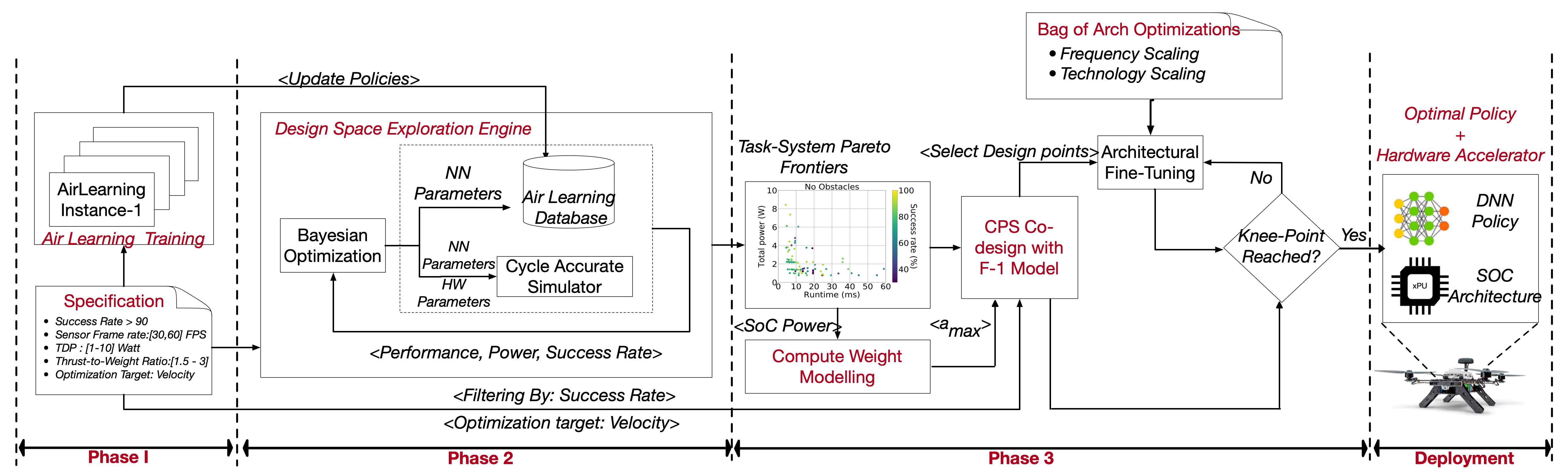}
  \caption{AutoPilot methodology for automating cyber-physical co-design in UAVs. \emph{Phase 1:} Multiple E2E models trained using Air Learning~\cite{airlearning} simulator based on high-level task specification. Success rates and hyper-parameters stored in a database. \emph{Phase 2:} Multi-objective DSE using Bayesian optimization to find E2E model and accelerator designs that are optimal in  success rate, and accelerator power and performance. \emph{Phase 3:} F-1 UAV tradeoff model~\cite{f-1} used to find the E2E model and accelerator that maximizes UAV mission performance.}
  \label{fig:autopilot}
\end{figure*}
\section{AutoPilot}
\label{sec:autopilot}
Autonomous UAV systems require many components for which there are many options to choose from, generating an enormous design space (\Fig{fig:uav-full-stack}).
To address the challenge of design space exploration over this combinatorial explosion of component choices, we introduce {\em AutoPilot}: an automation framework that intelligently navigates the large UAV design space and automatically generates a combination of hardware accelerator and autonomy algorithm for a given UAV type.

AutoPilot consist of three stages (see \Fig{fig:autopilot}). {\it Phase 1} prepares a collection of autonomy algorithm implementations that are functionally correct for performing autonomous UAV tasks. It takes an input specification of the autonomous UAV tasks and trains several end-to-end (E2E) autonomy algorithms for a given task using reinforcement learning (RL). {\it Phase 2} performs an automated design space exploration (DSE) using Bayesian optimization~\cite{bayesopt} to find the Pareto frontier of E2E algorithms and hardware accelerators that are optimal in terms of task success rate and power and runtime performance. {\it Phase 3} performs full-system UAV co-design based on its size, sensor characteristics (e.g., framerate and weight), and the design candidates (Pareto frontier designs) produced in Phase 2. Finally, a combination of the E2E algorithm and an accelerator is selected to maximize the number of missions.

\subsection{Phase 1: Task Specification and Training Using RL}
In this phase, the user provides  {\em task-level specification} for a UAV application (e.g., Source Seeking~\cite{source-seeking}) which includes some rough estimates about the environments such as obstacle densities. Based on this information, \autop configures the Air Learning~\cite{airlearning} environment generator. \autop uses Air Learning to train and validate E2E autonomy algorithms for a given UAV task. Air Learning provides a high-quality implementation of reinforcement learning algorithms that can be used to train a neural network navigation policy for the UAV. Air Learning includes a configurable environment generator~\cite{airlearning-github} with domain randomization~\cite{domain-rand1} that allows changing various parameters (e.g., the number of obstacles and size of the arena) to aid in generalizability.
These parameters are configured based on the autonomy task specification.



\begin{figure}[t]
\centering
  \subfloat[E2E model template.]{
	\begin{minipage}[c][0.25\columnwidth]{
	   0.5\columnwidth}
	   \centering
	   \includegraphics[width=1.0\textwidth]{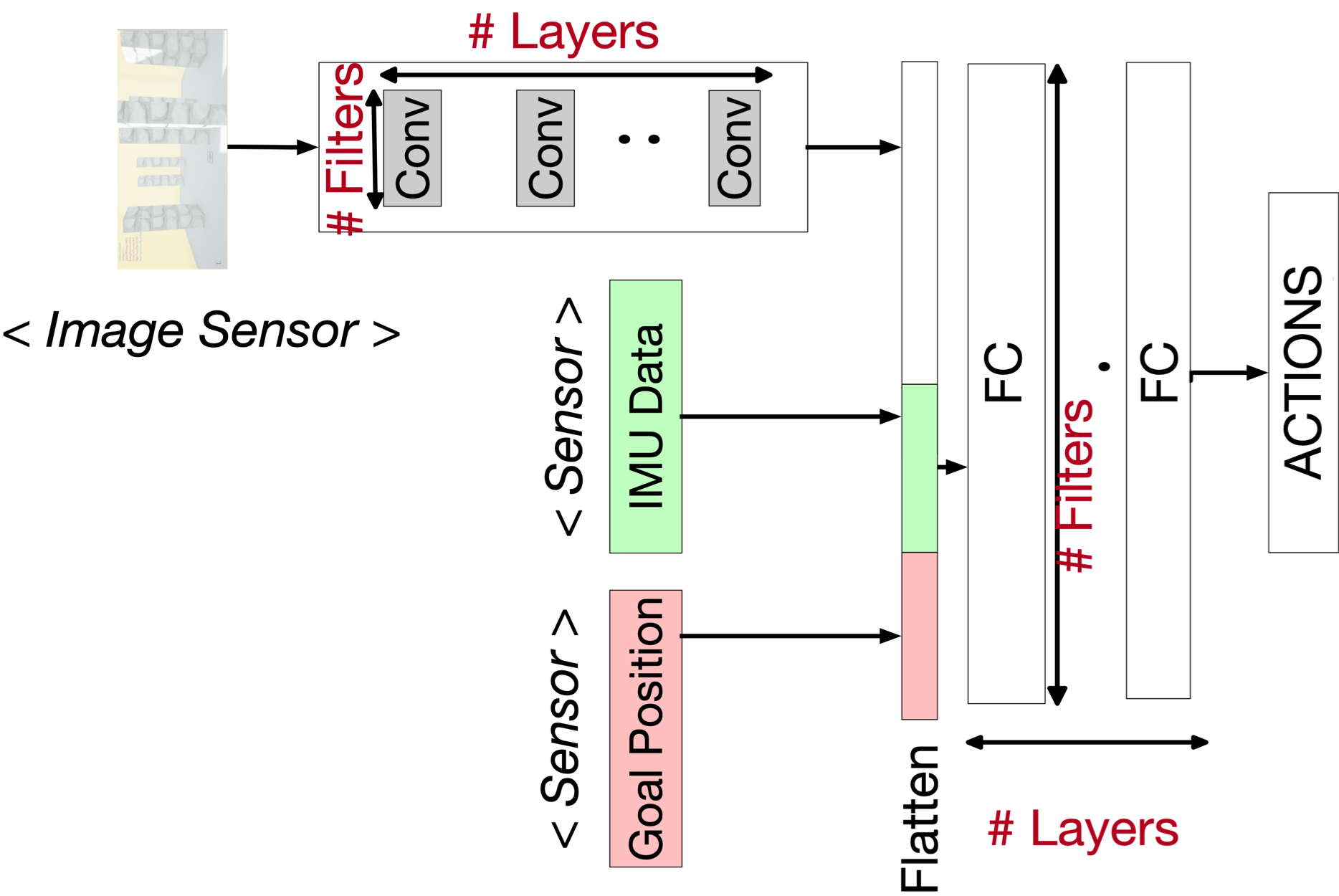}
	   \label{fig:e2e-template}
	\end{minipage}}
  \subfloat[Success Rate vs Model size.]{
	\begin{minipage}[c][0.25\columnwidth]{
	   0.5\columnwidth}
	   \centering
	   \includegraphics[width=1.0\textwidth]{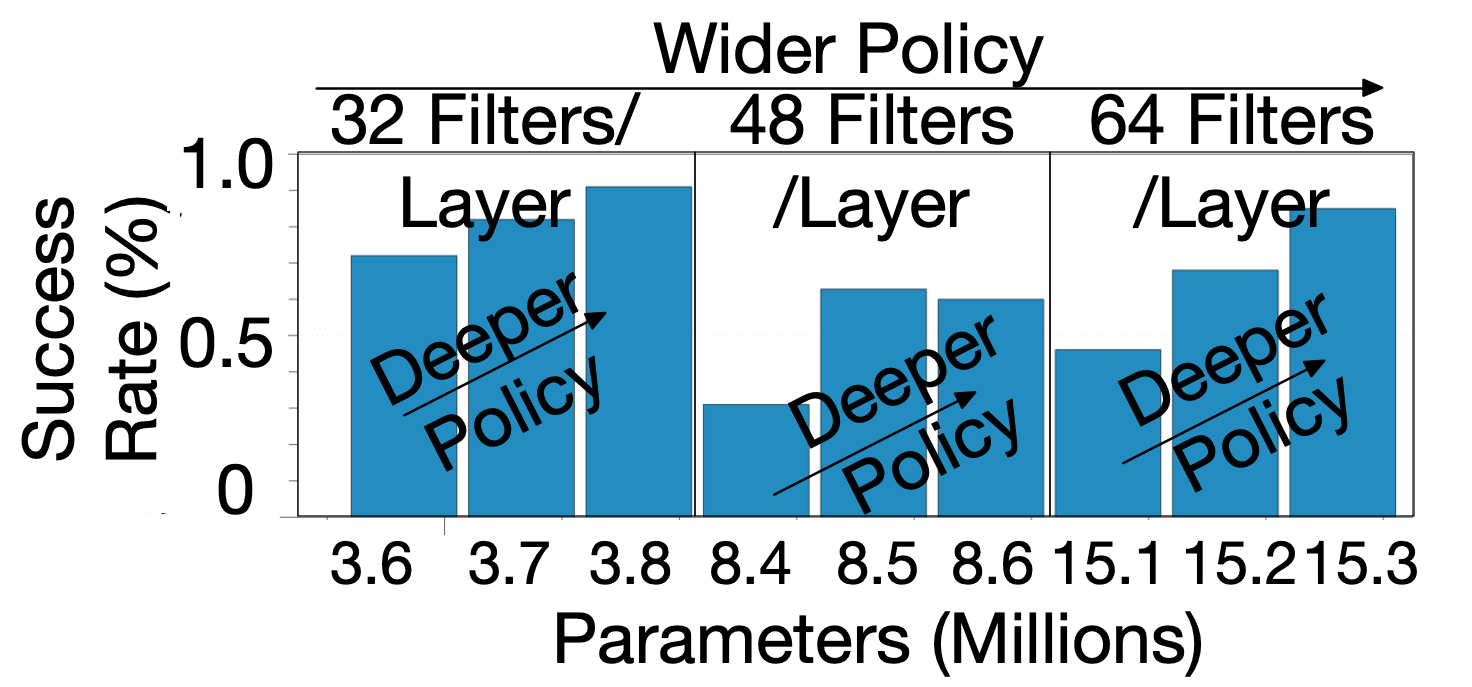}
	   \label{fig:params-task-trade-offs}
	\end{minipage}}
	
\caption{(a) Parameterized End-to-End (E2E) model template. (b) E2E models parameters vs. task-level success rate.}
\end{figure}

To determine the E2E model for each robot task (defined by environment complexity, i.e., obstacles), we start with the base template
used in Air Learning shown in \Fig{fig:e2e-template} and vary its hyperparameters (number of layers and filters) to create many candidate NN policies. We start from a known template and vary the parameters inside the template because not all the layers within the E2E model improve UAV task-level performance. Using domain knowledge, we can seed the search process to explore regions that quickly give us desired results.  For example, making the template layers deeper and wider gives a nice trade-off between the number of parameters and task-level success rate, as shown in \Fig{fig:params-task-trade-offs}. The task-level success rate of 60\% to 91\% is comparable to autonomous navigation task success rate reported in robotics literature~\cite{trailnet,guisti-trail,cad2rl,GUO2020,evdodgenet} for similar difficulty levels.

Based on these template parameters and the desired success rate, \autop launches several Air Learning training instances. Each of the NN policies that achieve the required success rate is evaluated in a domain random environment~\cite{domain-rand1}, and its task-level functionality is validated. The validated NN policies are updated into an Air Learning database along with their success rates, which are then used by Bayesian optimization in the next DSE phase (Section~\ref{sec:phase2}).

It is important to note that this NN parameter seed selection may be inappropriate for a different task (similar to how ImageNet trained models might fail if applied to medical images). The goal of this phase is to have the flexibility to train several E2E models, and there is no restriction on search space size. 

\subsection{Phase 2: Bayesian Optimization HW-SW Co-Design}
\label{sec:phase2}

In this phase, an automated multi-objective design space exploration is performed to find the Pareto frontier of E2E algorithms and hardware accelerator architectures that achieve optimal task success rate, performance, and power for a given autonomy task. The success rate is only affected by neural network hyperparameters (e.g., number of layers and filters). The accelerator's runtime and power depend on the E2E model and accelerator architectural parameters.

Success rates for the policies are accessed from the Air Learning database. At the same time, a cycle-accurate simulator is used to evaluate accelerator performance and power for the different policies and hardware configurations. Finally, we use Bayesian optimization to achieve rapid convergence to optimal solutions without performing an exhaustive search.

\textbf{Air Learning Database.}
This database stores the training results for the various E2E autonomy algorithms trained using Air Learning. Each entry in the database has an E2E algorithm identifier, the hyperparameters used for training, and the success rate for the policy after validation.

\begin{figure}[t]
\centering
  \subfloat[Hardware accelerator template.]{
	\begin{minipage}[c][0.4\columnwidth]{
	   0.5\columnwidth}
	   \centering
	   \includegraphics[width=1.0\textwidth]{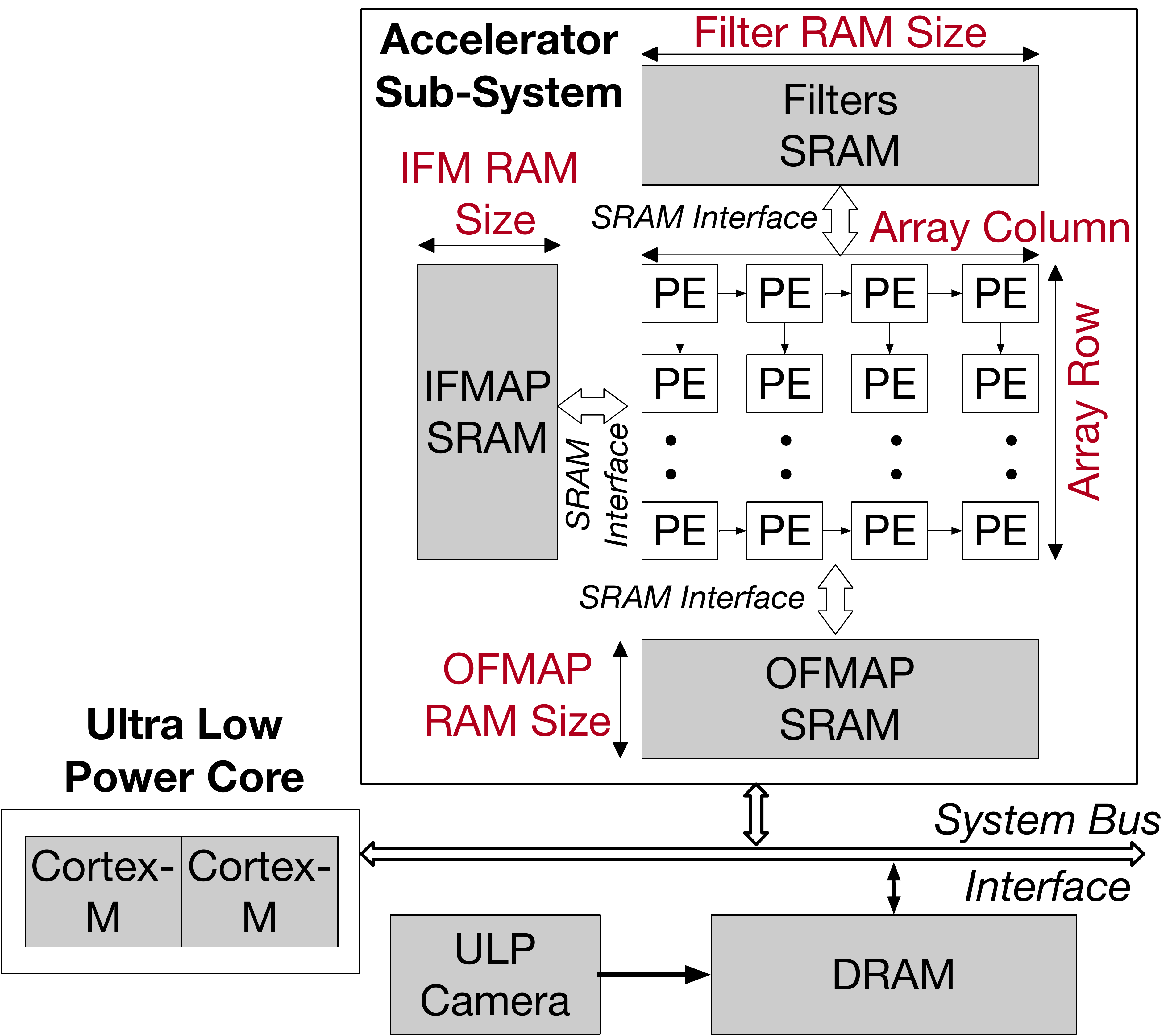}
	   \label{fig:soc-system}
	\end{minipage}}
  \subfloat[Performance and power trade-offs.]{
	\begin{minipage}[c][0.4\columnwidth]{
	   0.5\columnwidth}
	   \centering
	   \includegraphics[width=1.0\textwidth]{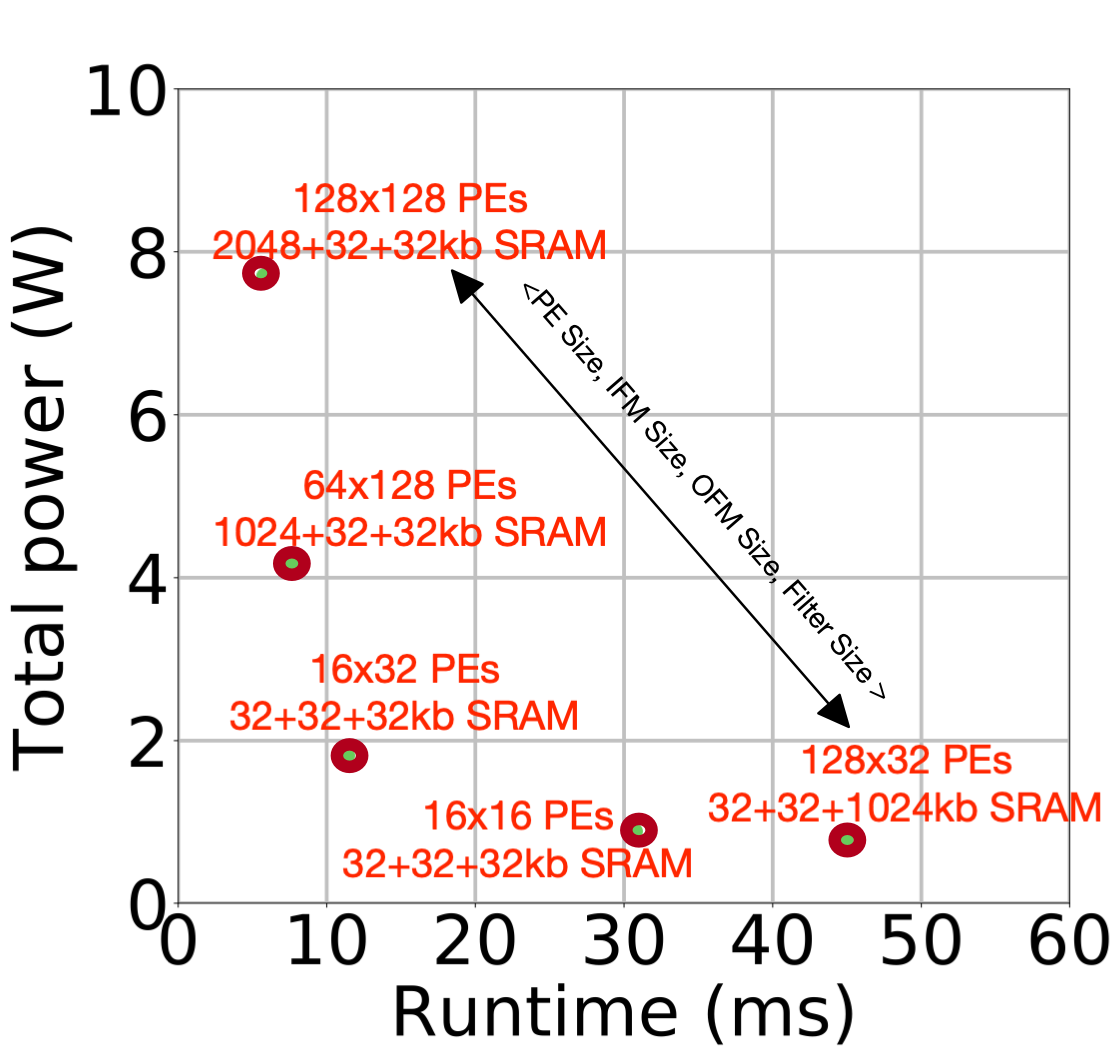}
	   \label{fig:perf-tradeoffs}
	\end{minipage}}
	
\caption{(a) Accelerator template containing parameterized systolic array template.(b) Varying of these parameters allows us to generate Pareto frontier designs.}
\end{figure}

\textbf{SoC Architecture.} We assume an SoC, which includes a parameterized template for hardware accelerator shown in \Fig{fig:soc-system}. The onboard compute system consists of two ultra-low-power cores for running the flight controller, an accelerator sub-system (Systolic array-based), an external memory (DRAM), and an onboard RGB sensor connected to the system bus. The flight controller software stack is a PID controller that runs bare-metal on the MCU, similar to Bitcraze Crazyflie aerial robots~\cite{crazyflie,source-seeking}. We also assume that the camera is interfaced with the system using a camera parallel interface~\cite{parallel} or MIPI~\cite{MIPI} similar to this work~\cite{pulp-dronet} from which the accelerator sub-system can directly fetch the inputs to process the images. In addition, we assume that the filter weights are loaded into the system memory as a one-time operation.

\textit{SoC Performance Estimation.} In the E2E autonomy paradigm, most of the time is spent processing the neural network~\cite{source-seeking,pulp-dronet,dronet}, and its performance will dominate the overall throughput of the autonomy algorithm. For each frame from the sensor, the E2E algorithm running on the accelerator sub-system generates high-level action commands that the flight controller interprets to generate low-level motor signals to control the UAV's physical rotors.

For determining accelerator performance for running a given E2E autonomy algorithm, \autop uses SCALE-Sim, a configurable cycle-accurate systolic array-based DNN accelerator simulator~\cite{samajdar2018scale}. In addition, it exposes various architectural parameters such as array size (number of MAC units), scratchpad memory size for the input feature maps (ifmap), filters and output feature maps (ofmaps), dataflow mapping strategies, as well as system integration parameters, e.g., memory bandwidth. For example, enumerating \# PE's, SRAM sizes give a nice trade-off between performance and power as shown in \Fig{fig:perf-tradeoffs}. Taking these architectural parameters, the SCALE-Sim generates the runtime latency.

\textit{SoC Power Estimation.} For estimating the total SoC power, we add the power of individual components in the SoC. For estimating the power of the accelerator, we run a given NN policy on a cycle-accurate simulator. The cycle-accurate simulator produces SRAM traces, DRAM traces, number
of read/write access to SRAM, number of read/write access to the DRAM. Using the SRAM and DRAM trace information, we model the SRAM power in CACTI~\cite{cacti} and DRAM power in Micron DRAM model~\cite{dram}. For estimating the power for the systolic array, we multiply the array size with the energy of the PE. The PE power is modeled after the breakdown in \cite{li-memdse-dac}.

For the MCU cores, we use Cortex-M cores that implement the ARMv8-M ISA~\cite{arm-v8m-isa}.
Each core consumes about 0.38 mW in a 28 nm process clocked at 100 MHz~\cite{arm-m33-productpage}. We also account for the power of the ultra-low-power core into our final power numbers. For the ULP camera, we assume it consumes 100 mW and form factor of 6.24 mm $\times$ 3.84 mm~\cite{ulp-camera}. We account for the camera power in our overall power calculation.





\textbf{Bayesian Optimization.}
\autop uses Bayesian optimization~\cite{bayesopt} for algorithm-HW design space exploration. Bayesian optimization has shown to be effective for optimizing black-box functions~\cite{SnoekLA12,ShahriariSWAF16} that are expensive to evaluate and cannot be expressed as closed-form expressions.

In \autop, we use bayesian optimization to optimize three objective functions: (i) task success rate, (ii) SoC power, and (iii) accelerator inference latency (runtime). A Pareto-optimal design achieves maximum task success rate, minimum inference latency, and SoC power. The algorithm tunes NN policy hyper-parameters (such as number of layers and filters) and accelerator hardware parameters (e.g., number of processing elements, SRAM sizes) to converge to Pareto-optimal NN policies and accelerator architectures. 

We use an open-source BayesOpt implementation~\cite{bayesopt} is used in \autop. BayesOpt initially evaluates the objective functions at random parameters, followed by intelligently selecting those that will optimize the objectives. The algorithm builds a Bayesian statistical model for each objective function: a Gaussian process (GP) is used. These GP models are updated as the BayesOpt proceeds and samples new parameters. A GP distribution is defined by a mean and covariance. The mean is the expected value of a function at some parameter value. The covariance, called the kernel, models the dependence between the function values at two distinct parameter values. In this paper, the widely-used squared exponential (SE) kernel is used due to its simplicity, leading to fast computation~\cite{gp}. 
The right selection of parameter values is determined by an acquisition function computed using the GP-predicted objective values. The algorithm selects those inputs that maximize the acquisition function until all the optimal solutions are found. 

In particular, the {\em S-Metric-Selection-based Efficient Global Optimization (SMS-EGO)} is used as the acquisition function. It has been shown to be highly effective for multi-objective optimization and handling a large design space compared to other acquisition strategies such as expected improvement~\cite{smsego}. SMS-EGO uses {\em hypervolume} to determine the degree to which a candidate point is optimal: it is the volume enclosed between a candidate point and a fixed reference point in the Pareto space. Since the hypervolume of a Pareto-optimal point is higher than a non-optimal point, the approach maximizes the hypervolume until all the points with the highest hypervolume (i.e, Pareto-optimal) are found.

However, not all Pareto-optimal compute designs generated at this stage will result in optimal UAV performance (e.g., maximize the number of missions). Some of these designs will be over-provisioned or under-provisioned. Both these scenarios negatively affect the overall UAV performance. Hence, to determine which of these designs is better suited for a given UAV, we need to performance full-system UAV co-design where we account for sensor, onboard compute performance, and its impact on UAV physics.

\subsection{Phase 3: Cyber-Physical System Co-Design}
\label{sec:cps-co-design}
\autop's Phase 2 prunes a large design space of $\approx$10\textsuperscript{18} designs to $\approx$100s of design candidates. These 100s of design candidates represent a sample of low-power, high-performance, or Pareto-optimal designs in terms of performance and power. However, enumerating 100s of design candidates manually is still tedious and requires a systematic way of selecting one of the designs for a UAV. The goal of Phase 3 is to determine the holistic evaluation of these design candidates and other UAV components.



\begin{figure}[t]
  \subfloat[Effect of compute weight.]{
	\begin{minipage}[c][0.2\columnwidth]{
	   0.55\columnwidth}
	   \centering
	   \includegraphics[width=1.0\textwidth]{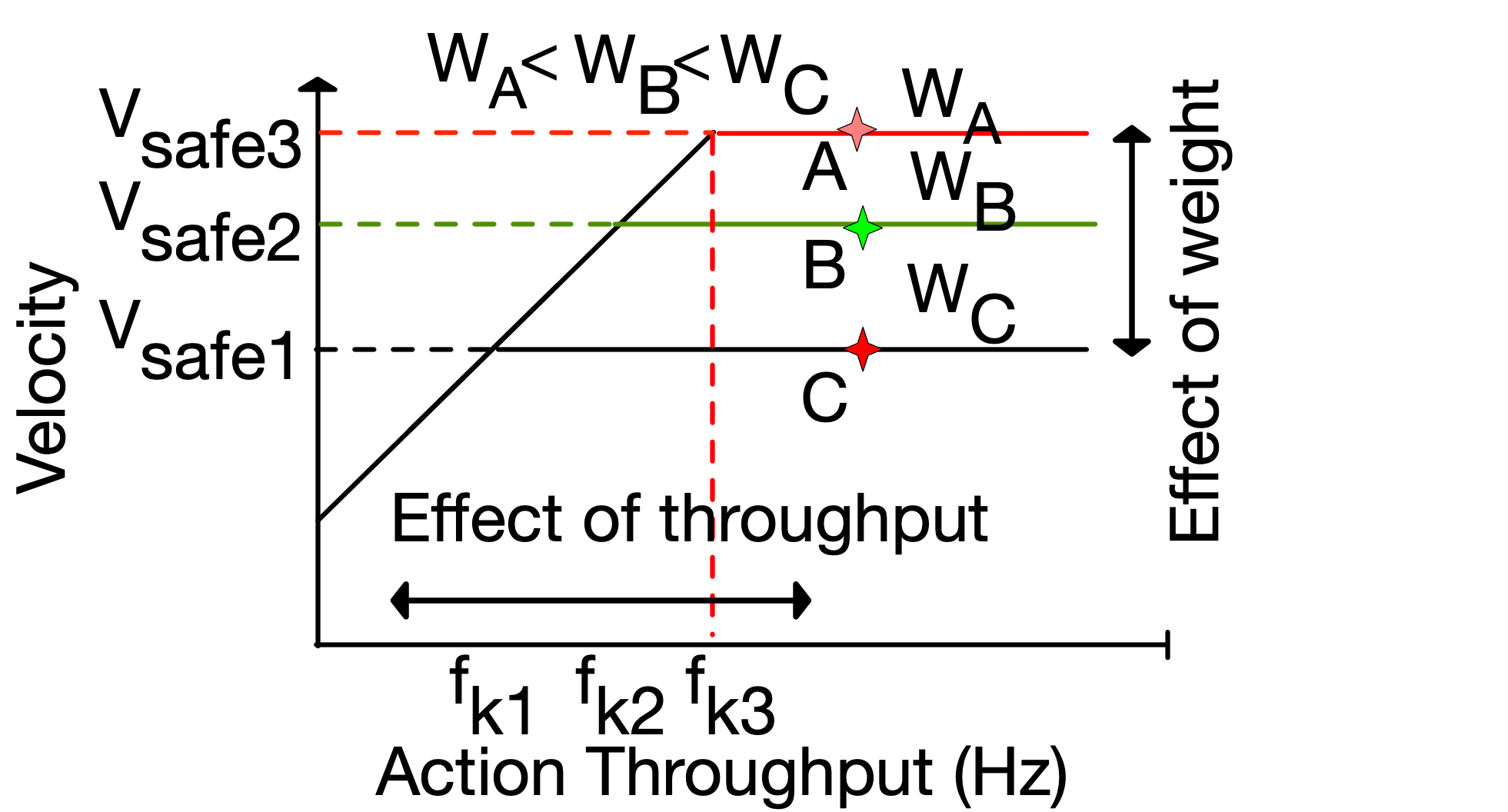}
	   \label{fig:f1-effects}
	\end{minipage}}
 \hspace{1pt} 	
  \subfloat[Optimal design.]{
	\begin{minipage}[c][0.2\columnwidth]{
	   0.4\columnwidth}
	   \centering
	   \includegraphics[width=1.0\textwidth]{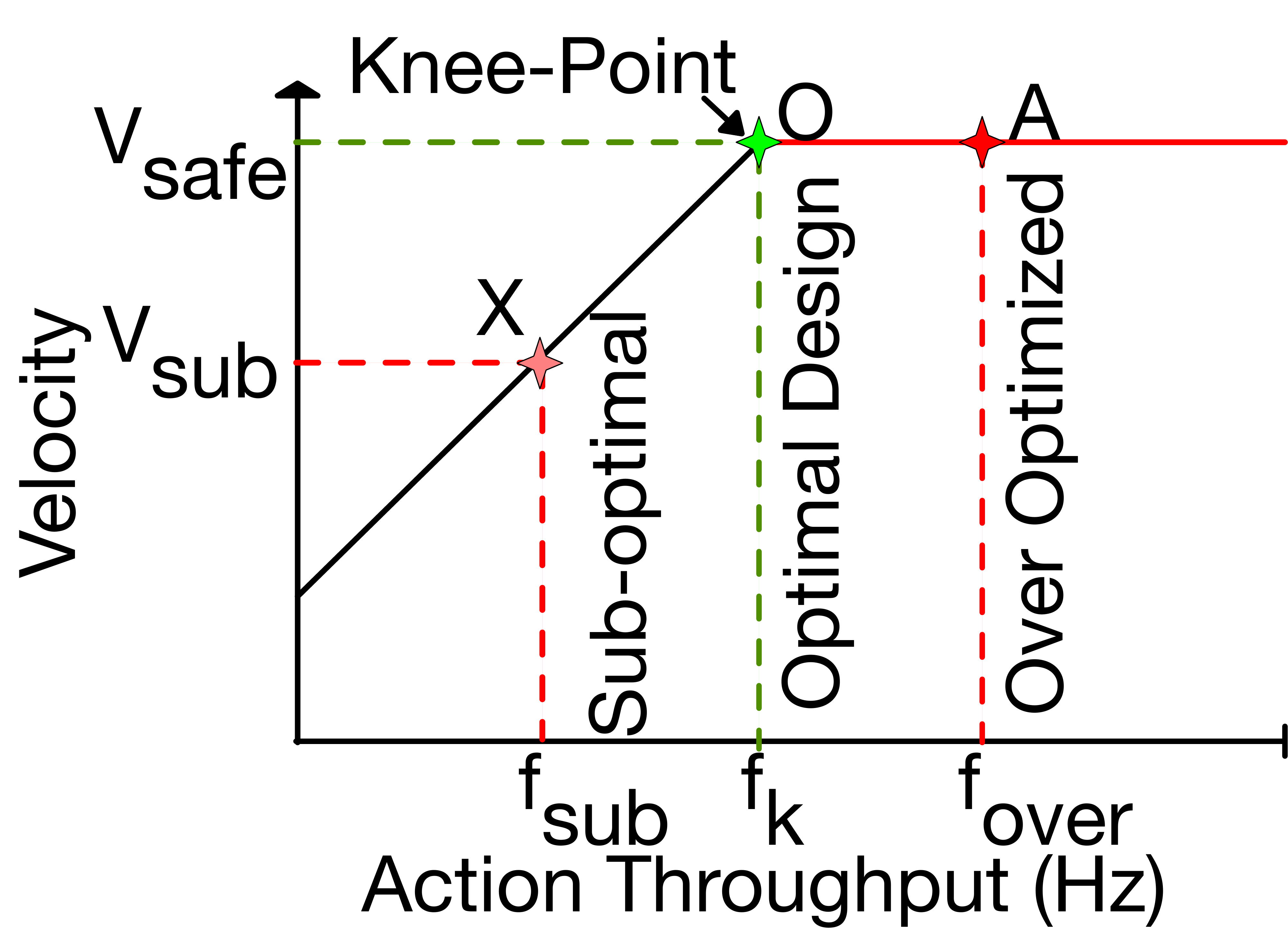}
	   \label{fig:optimality}
	\end{minipage}}
\caption{(a) Mapping design candidates to F-1 to assess impact of compute weight and compute throughput on UAV performance. (b) Mapping design candidates to F-1 to select optimal design candidate for a UAV system.\vspace{-10pt}}
\end{figure}

\textbf{Compute Weight Modelling.} 
Since the payload weight affects the UAV physics, it is important to estimate the weight of the onboard compute. The onboard compute typically has two components: a motherboard where SoC is mounted and other electrical components and a passive heatsink for cooling the SoC. The heatsink weight is proportional to the TDP. 

For the motherboard weight, we assume that the final SoC is mounted on a PCB along with all electrical components weighing 20g (which per our analysis is typical for Ras-Pi~\cite{ras-pi-weight}, CORAL~\cite{coral-weight} like systems).
For the heatsink weight, we use a heat sink calculator~\cite{heat-sink} which determines the heatsink volume required for cooling. The weight of the heatsink is then determined by multiplying the estimated volume with the density of aluminum (commonly used heatsink material).

{\bf Cyber-Physical Co-Design Using F-1 Model.} First, designs with the highest success rate (based on the input specification) are filtered from the designs generated in Phase 2. 
Next, all the filtered design candidates are mapped to the F-1 UAV tradeoff model~\cite{f-1}. Each base UAV system has a unique F-1 plot that can gather insights about different bounds and bottlenecks. F-1~\cite{f-1} model is a roofline-like visual performance model built on top of a safety model~\cite{high-speed-drone} for UAVs. 

The F-1 model plots the relationship between how fast but safely a UAV can travel (safe velocity) based on its decision-making rate (output of sensor-compute-control pipeline). Conversely, by relative motion (and switching the frame of reference), if the UAV is hovering, the same model can also tell the maximum velocity of an incoming object the UAV can avoid before colliding.
In addition, the model takes into account the compute throughput and considers how the payload weight affects the UAV's physics--i.e., maximum acceleration (which can be determined by its thrust-to-weight ratio~\cite{twr-amax}). Another important insight it can provide is whether a combination of autonomy algorithm and onboard compute is over-provisioned, under-provisioned, or optimal for a given UAV system.

To intuitively understand the role of F-1 in cyber-physical system co-design, let us consider three hypothetical designs, `A,' `B,' and 'C' generated from Phase 2. Let us also assume that all the designs achieve the same compute throughput but at different power (TDPs), with `A' being lowest and `C' being the highest. Mapping these designs in the F-1 model (\Fig{fig:f1-effects}) can give insights into which of these three designs will achieve better UAV performance. Since `A' has the same compute performance with the lowest power, it will also have the lowest heatsink weight, while `C' has the highest power, it will have the highest heatsink weight. The weight of the onboard compute affects UAV ability to move faster, and the lowering of ceilings captures this effect in \Fig{fig:f1-effects} for `B' and `C.' Lowering of safe velocity will have an implication on mission energy and the number of missions it can perform~\cite{mavbench}. Thus, \autop will select `A' over the other two designs.

 \begin{figure*}[t]
  {\centering
	\begin{minipage}[c][0.1\columnwidth]{
	  1.85\columnwidth}
	   \centering
	   \includegraphics[width=1.0\textwidth]{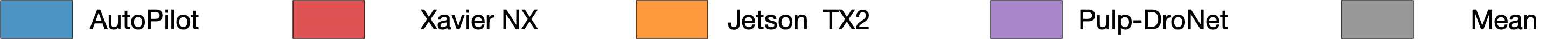}
	\end{minipage}}
\subfloat[Low-obstacle: Number of missions.]{
	\begin{minipage}[c][0.3\columnwidth]{
	   0.66\columnwidth}
	   \centering
	   \includegraphics[width=1.0\textwidth]{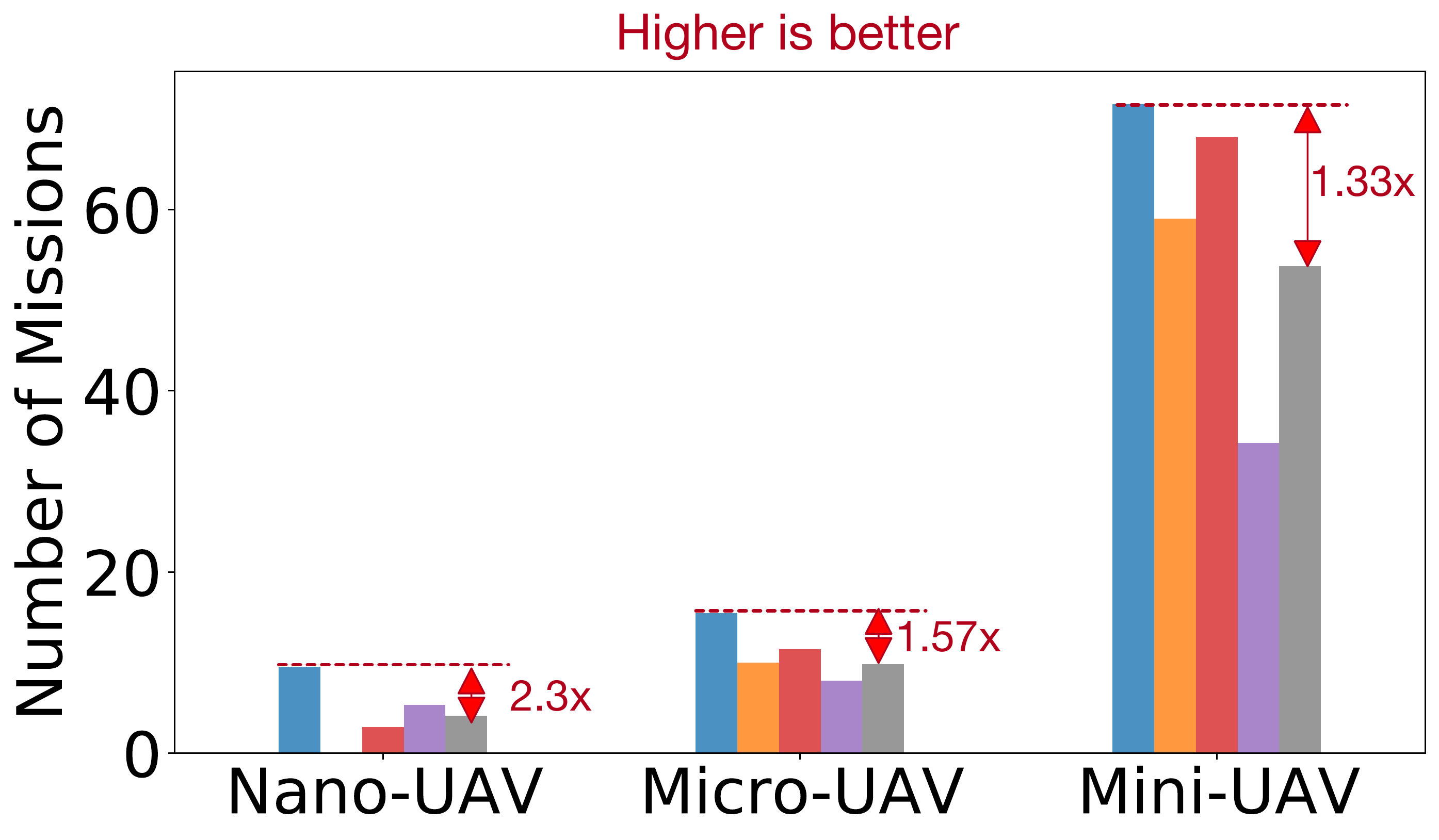}
	   \label{fig:mission-sorties-qos-low}
	\end{minipage}} 
\subfloat[Some-obstacle: Number of missions.]{
	\begin{minipage}[c][0.3\columnwidth]{
	   0.66\columnwidth}
	   \centering
	   \includegraphics[width=1.0\textwidth]{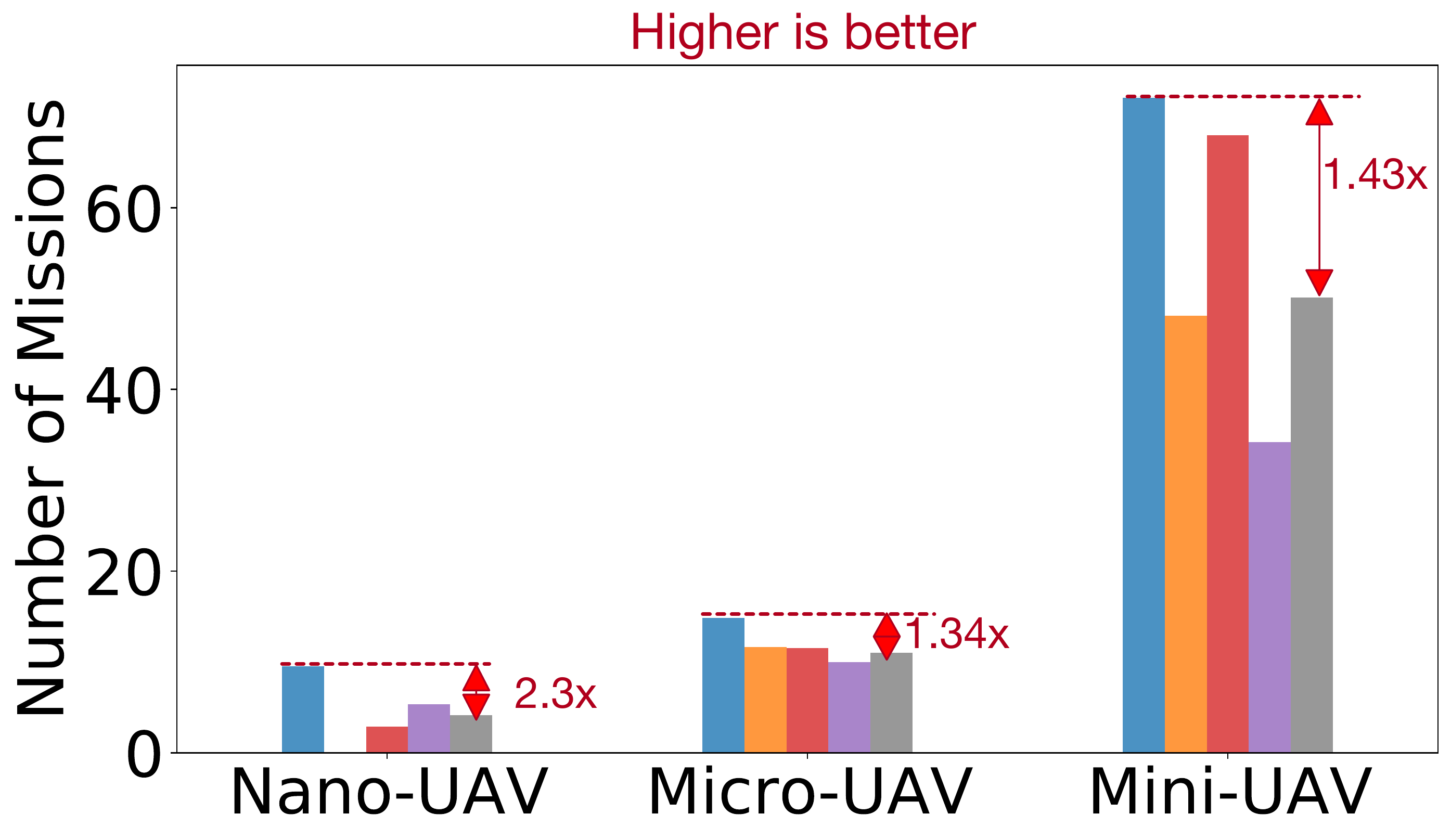}
	  \label{fig:mission-sorties-qos-med}
	\end{minipage}} 
\subfloat[Dense-obstacle: Number of missions.]{
	\begin{minipage}[c][0.3\columnwidth]{
	   0.66\columnwidth}
	   \centering
	   \includegraphics[width=1.0\textwidth]{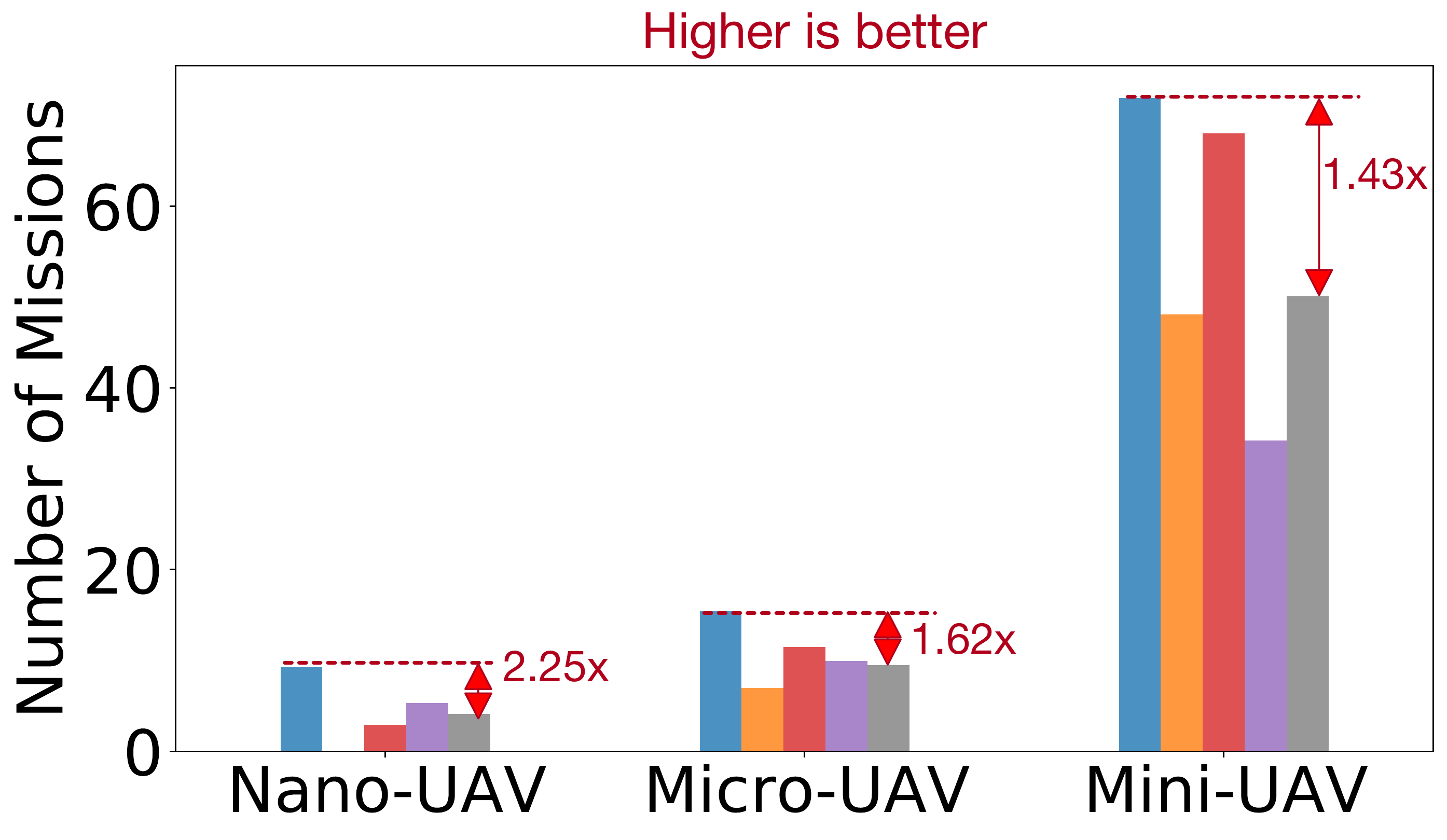}
	  \label{fig:mission-sorties-qos-den}
	\end{minipage}}
\caption{Comparison of AutoPilot generated designs with other designs (TX2 and Xavier NX, PULP~\cite{pulp-dronet}) for three different deployment scenarios (low obstacle, medium obstacles, and dense obstacle environments) and three drone categories (mini-UAV, micro-UAV, and nano-UAVs). (a), (b), and (c) denotes the mission the number of missions \textit{(higher is better)} on a single battery charge for three drone types across three different deployment scenario. The degradation in the \# of missions are compared with mean performance of TX2, NX, and PULP~\cite{pulp-dronet} and annotated in the plots with \autop generated design as baseline. All the points except P-DroNet (PULP-DroNet)~\cite{pulp-dronet} runs the same policy. For P-DroNet, we use the numbers reported from their work.}
\label{fig:mission-qos}
\end{figure*}

Another utility of the F-1 model in \autop is determining if a design candidate is optimal, over-provisioned, or under-provisioned. The minimum value of action throughput to maximize safe velocity (V$_{safe}$) is called the knee-point. For example, in \Fig{fig:optimality}, design `O' is optimal because it achieves the minimum action throughput to maximize V$_{safe}$. Likewise, `A' is over-provisioned (achieves more throughput than required), and `X' is under-provisioned. \autop will choose design `O' over the other two designs.



{\bf Architectural Fine-Tuning.} In the case when no optimal design exists that achieves the knee-point, some architectural tuning may be required to shift the design close to the knee-point. \autop provides two options for which points to consider for optimization: (i) these can be user-defined, or (ii) the design point closest to the knee-point can be selected. Architectural tuning can be performed using various optimizations until the optimized design is at (or very close to) the base knee-point in the F-1 roofline. To this end, we employ a ``bag'' of architectural optimizations in the tuning process and seed \autop with two techniques: frequency scaling and technology scaling.


\section{Experimental Setup}
\label{sec:eval}

This section describes the details of our setup to evaluate the efficacy of \autop in generalizing onboard compute design across many UAV types. We also provide details about the UAV types, autonomy tasks, and sensor information used to perform full-UAV system co-design.



\subsection{Autonomy Task}
\label{sec:autonomy_tasks}
We train autonomous navigation tasks in Air Learning~\cite{airlearning} for three different environments: low, medium, and dense-obstacles. 
Point-to-point navigation is one of the key building blocks in achieving autonomy and is used in many practical applications like search and rescue, source-seeking~\cite{source-seeking}, and package delivery.
In the low-obstacle scenario, there are four obstacles with goal position randomly changing every training episode.
In the medium scenario, there are four fixed obstacles and up to three randomly-placed obstacles.
In the dense obstacle scenario, there are four fixed obstacles in the dense scenario and up to five randomly placed obstacles. Each E2E model is trained for 1~Million steps or until convergence. This is a standard training methodology for reinforcement learning, and the same methodology is used for building real autonomous UAV applications~\cite{source-seeking,cad2rl}.

\subsection{Base UAV Systems and Autonomy Components}

To show the scalability of \autop methodology, we take one representative UAV from each size category: mini, micro, and nano-UAV (Table~\ref{tab:uav-spec}).  The base UAV system includes a frame, flight controller, battery, and rotors (all included in the base weight). We keep the base UAV system fixed, and focus on designing the optimal onboard compute and autonomy algorithm to maximize the overall operational efficiency of the autonomous UAV system. 

\subsection{Autonomous UAV Evaluation Metrics}

An important operational efficiency metric for autonomous drones is the `\textit{number of missions}', which captures how many times the drone can complete similar missions on a single battery charge. 
For example, in a package delivery use case, a higher number of missions means more packages delivered with lower downtime spent recharging.
This metric is affected by the choice of drone onboard compute combined with several other key components.

For a given drone, we define the number of missions as:
\begin{equation}
N_{missions} = \frac{E_{battery}}{E_{mission}}
\label{eq:num-missions}
\end{equation}
where $E_{battery}$ is the total energy available in the drone (a function of battery mAh rating) and $E_{mission}$ is the total energy expended by the drone per mission.

We can define $E_{mission}$ for a single mission as:
\begin{equation}
 E_{mission} =  (P_{rotors} + P_{compute} + P_{others})* t_{mission}
\label{eq:mission-energy}
\end{equation}
where $P_{rotors}$, $P_{compute}$, and $P_{others}$ refer to the power consumption of the rotor propulsion, compute, and other electronic components (e.g., sensors, ESC) in the drone. $t_{mission}$ is the time for completing the mission. Intuitively, Eq~\ref{eq:mission-energy} suggests that the amount of energy expended in a mission corresponds to the duration of the time the drone flies (mission time), and the total power dissipation of its components.

The mission time $t_{mission}$ depends upon the distance $D_{operation}$ and the UAV velocity. For a fixed distance, mission time is determined by how fast the drone can travel through a dynamic environment filled with obstacles. The drone needs to travel as quickly as possible while safely navigating around obstacles. We define this safe traveling speed as the safe velocity, $V_{safe}$. Maximizing $V_{safe}$ lowers the mission time, increasing the total possible number of missions.

Using these terms, we can re-write Eq~\ref{eq:mission-energy} as follows:
\begin{equation}
 E_{mission} =  (P_{rotors} + P_{compute} + P_{others})* \frac{D_{operation}}{V_{safe}},
 \label{eq:mission-energy-expand}
\end{equation}
Then, substituting Eq~\ref{eq:mission-energy-expand} in Eq~\ref{eq:num-missions}, we get:
\begin{equation}
N_{missions} = \frac{E_{battery}* V_{safe}}{(P_{rotors} + P_{compute} + P_{others})*D_{operation}}
 \label{eq:num-missions-expanded},
\end{equation}

According to Eq.~\ref{eq:num-missions-expanded}, to maximize the number of missions, the optimization objective is to increase the UAV's safe velocity  ($V_{safe}$) or increase the battery capacity ($E_{battery}$).  Increasing the battery capacity is non-trivial since drone size impacts the SWaP constraints. However, proper selection of various UAV components (compute, sensors, etc.) can maximize safe velocity. It is important to note that $E_{mission}$, $V_{safe}$, and $D_{operation}$ are not constants and it depends upon the mission characteristics. But the fundamental relationship between them holds true.

\begin{table}[t]
\renewcommand\arraystretch{1.2}
\resizebox{1.0\columnwidth}{!}{
\begin{tabular}{|c|c|c|c|c|c|c|c|c|}
\hline
\multirow{2}{*}{\textbf{\begin{tabular}[c]{@{}c@{}}Drone\\ Name\end{tabular}}} & \multicolumn{5}{c|}{\textbf{\begin{tabular}[c]{@{}c@{}}Base-UAV System \\ (Fixed)\end{tabular}}} & \multicolumn{3}{c|}{\textbf{\begin{tabular}[c]{@{}c@{}}Autonomy Components\\  (Custom Designed)\end{tabular}}} \\ \cline{2-9} 
 & \textbf{\begin{tabular}[c]{@{}c@{}}UAV \\ Type\end{tabular}} & \textbf{\begin{tabular}[c]{@{}c@{}}Battery\\ Capacity\\ (mAh)\end{tabular}} & \textbf{\begin{tabular}[c]{@{}c@{}}Flight\\ Controller\end{tabular}} & \textbf{\begin{tabular}[c]{@{}c@{}}Base \\ UAV\\ Weight\end{tabular}} & \textbf{Sensor} & \textbf{\begin{tabular}[c]{@{}c@{}}Sensor\\ Framerate\end{tabular}} & \textbf{\begin{tabular}[c]{@{}c@{}}Autonomy\\ Algorithm\end{tabular}} & \textbf{\begin{tabular}[c]{@{}c@{}}Onboard\\ Compute\end{tabular}} \\ \hline
\textit{\textbf{AscTec Pelican}} & \textit{mini-UAV} & \textit{\begin{tabular}[c]{@{}c@{}}6250\\ (fixed)\end{tabular}} & \textit{\begin{tabular}[c]{@{}c@{}}PID \\ Controller\\ 100 KHz\end{tabular}} & \textit{1650 g} & \textit{RGB} & \textit{30/60 FPS} & \textit{\begin{tabular}[c]{@{}c@{}}E2E\\ (custom)\end{tabular}} & \textit{Custom} \\ \hline
\textit{\textbf{DJI Spark}} & \textit{micro-UAV} & \textit{\begin{tabular}[c]{@{}c@{}}1480\\ (fixed)\end{tabular}} & \textit{\begin{tabular}[c]{@{}c@{}}PID\\  Controller\\ 100 KHz\end{tabular}} & \textit{300 g} & \textit{RGB} & \textit{30/60 FPS} & \textit{\begin{tabular}[c]{@{}c@{}}E2E\\ (custom)\end{tabular}} & \textit{Custom} \\ \hline
\textit{\textbf{Zhang et al~\cite{nano-UAV}}} & \textit{nano-UAV} & \textit{\begin{tabular}[c]{@{}c@{}}500\\ (fixed)\end{tabular}} & \textit{\begin{tabular}[c]{@{}c@{}}PID \\ Controller\\ 100 KHz\end{tabular}} & \textit{50 g} & \textit{RGB} & \textit{30/60 FPS} & \textit{\begin{tabular}[c]{@{}c@{}}E2E\\ (custom)\end{tabular}} & \textit{Custom} \\ \hline
\end{tabular}}
\caption{In our experiments, we keep the base UAV system fixed (size, battery, sensor type) and focus on co-design of components needed for achieving autonomy.\vspace{-10pt}}
\label{tab:uav-spec}
\end{table}

\section{Evaluation}
\label{sec:results}
This section evaluates \autop's co-design results for different UAVs and deployment scenarios compared to baseline designs.
We then compare \autop methodology in selecting designs versus several traditional design strategies. Next, we characterize the effects of cyber-physical parameters such as sensor type and UAV agility on the compute co-design process. Finally, we analyze the cost of design specialization versus its impact on overall UAV mission efficiency.


\subsection{Co-Design for Different UAVs and Environments}
\label{sec:scalability}
In this experiment, we demonstrate that the AutoPilot methodology is \emph{scalable and generalizable} in generating designs that maximize the number of missions across three UAV types and three different deployment environment scenarios.

The \emph{different classes of drones} evaluated are a mini-UAV (AscTec Pelican), a micro-UAV (DJI-Spark), and a nano-UAV (used in Zhang et al.~\cite{nano-UAV}) whose specifications are in Table~\ref{tab:uav-spec}.
The \emph{different deployment scenarios} with varying levels of complexity evaluated are: low, medium, and dense-obstacles (details in Section~\ref{sec:autonomy_tasks}). These auto-generated domain randomized environments have varying degrees of obstacle densities. They represent common deployment use cases for drones. For instance, autonomous navigation for a farming use case could be very sparse (low-obstacle), whereas a search and rescue operation in a forest is a dense-obstacle scenario.

The co-design goal is choosing an onboard compute system and autonomy algorithm that minimizes the mission time and mission energy, maximizing the number of missions the UAV can perform on a single battery charge (Eq.~\ref{eq:num-missions} and Eq.~\ref{eq:mission-energy-expand}).

\textbf{Co-Design Comparison Results.} \Fig{fig:mission-qos} shows the comparison in the number of missions between design generated using \autop methodology and two general-purpose selections (Jetson TX2 and Xavier NX) and PULP~\cite{pulp-dronet}. On a fully charged battery, \autop generated optimal designs on average achieves 1.33$\times$ to 1.43$\times$ more missions for mini-UAV compared to TX2, Xavier NX, and PULP~\cite{pulp-dronet}. For micro-UAVs, \autop on average achieves 1.34$\times$ to 1.62$\times$ . Likewise, for nano-UAV, \autop on average achieves  2.3$\times$ more missions, thus enabling higher operational efficiency than an ad-hoc selection of general-purpose designs. 

The \autop methodology for the generation of onboard hardware and autonomy algorithm for these UAV types and deployment scenarios consistently outperforms general-purpose hardware (Jetson TX2 and Xavier NX) and a domain-specific hardware accelerator built for UAVs~\cite{pulp-dronet}.
This demonstrates the flexibility of \autop, compared to prior works that focus on a specific UAV type~\cite{aspolos-drone,navion,source-seeking,pulp-dronet}.

\begin{figure}
\vspace{5pt}
        \includegraphics[width=1.0\columnwidth]{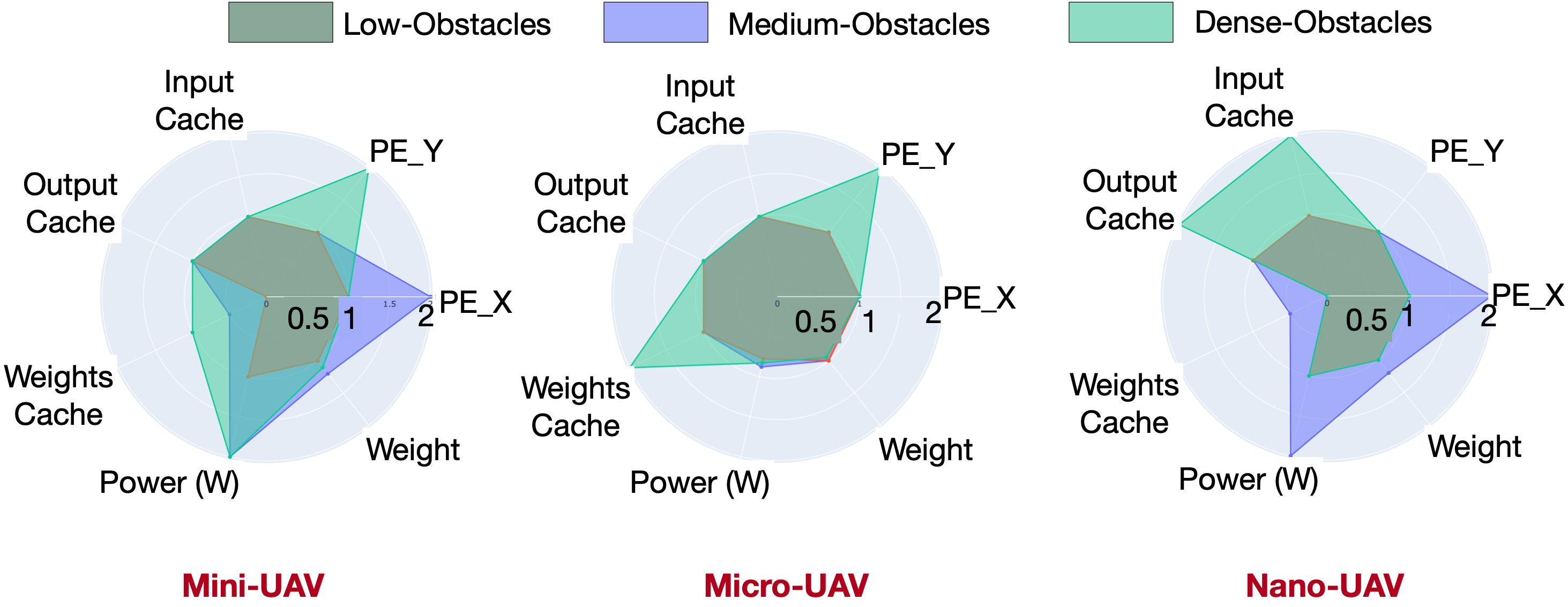}
        \caption{Visualization of architectural parameter variations for nine scenarios (three UAVs and three deployment scenarios). The scales are normalized with respect the lowest value for each architectural parameter to show the variations requirements as the UAV components changes.}
        \label{fig:arch-params-variations}
\end{figure}

\textbf{Analysis of Architectural Parameter Variations.} Visualization of various architectural parameters selections across all the nine scenarios (three UAVs and three deployment scenarios) to gain insights into the designs that \autop generates is shown in \Fig{fig:arch-params-variations}.  Deployment scenario affects the  complexity of the E2E models. A model with five layers and 32 filters achieves the highest success rate for the low obstacle environment. For the dense environment, the model with seven layers and 32 filters achieves the highest success rate.

\noindent The increasing complexity of the autonomy algorithm means the onboard compute for a specific UAV has to achieve similar performance while keeping the power and weight constant. Using Bayesian optimization allows \autop to select architectural parameters (PE size, cache size, etc.) intelligently to satisfy the performance, power, and weight constraints for the mini-UAV as shown in \Fig{fig:arch-params-variations}. For instance, as the complexity of the task increases, the compute design becomes bigger (as seen by the larger spread in parameter space \Fig{fig:arch-params-variations}). But \autop only chooses those parameters to keep the power and weight nearly the same (or never exceed a certain limit). Manually enumerating these points (isolated or one-size accelerator designs) would not maximize the mission performance as the UAV type or deployment scenario changes. Thus, automating the co-design space intelligently allows \autop to consistently generate designs that maximize mission performance.

\begin{figure}

\vspace{5pt}
        \includegraphics[height=0.45\columnwidth,width=1.0\columnwidth,keepaspectratio]{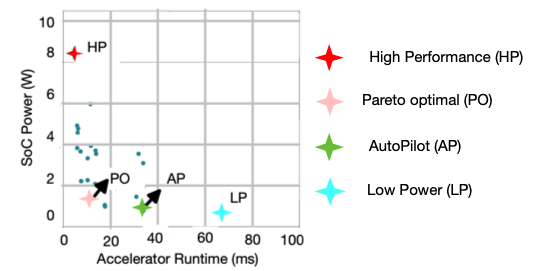}
        \caption{ Phase 2 Pareto frontier designs with different performance/power profiles. We highlight four candidates (annotated as HP-high performance, LP-low power, PO-Pareto optimal in terms of performance/power, and AP-Autopilot).\vspace{-2pt}}
        \label{fig:need-cps}
\end{figure}

\subsection{Comparison to Traditional Design Strategies}
\label{sec:cps-codesign}

In the following experiments, we demonstrate the importance of performing a full-system UAV co-design rather than selecting onboard compute based on traditional design strategies: maximizing compute performance alone; optimizing for low power alone; and even Pareto optimal selection with respect to both compute performance and power efficiency.
Our results show that performing full-system UAV co-design is a necessity for maximizing overall mission efficiency.

Ideally, what differentiates between full-system UAV co-design versus methodologies outlined in prior work~\cite{ReagenHAGWWB17} is the critical Phase 3 in \autop (\Fig{fig:autopilot}). Hence, we take an intermediate output from Phase 2 of \autop methodology and show how a lack of full-system co-design degrades the mission performance of autonomous UAVs. To demonstrate the degradation concretely, we choose a nano-UAV whose specifications are available in Table~\ref{tab:uav-spec}. The output of Phase 2 (from \Fig{fig:autopilot}) for this task are shown in \Fig{fig:need-cps}.


Based on these strategies outlined above, we label these designs in \Fig{fig:need-cps} as `HP' (high-performance design), `LP' (low-power design), and `PO' (Pareto optimal design). We also label another design that does not use the conventional selection strategies and instead uses \autop methodology. We label this point as `AP' (\autop selected design). Finally, we compare the mission level performance of `AP' and others to gain the following architectural insights.

\subsubsection{Comparison to Optimizing for Performance Alone}

We compare the high-performance design (HP) with \autop generated design (AP) (\Fig{fig:need-cps}) to show that \emph{high performance compute design does not directly translate to overall high autonomous UAV performance}. HP achieves a compute throughput of 205 FPS while consuming 8.24 W (65 g) whereas AP achieves a throughput of 46 FPS at 0.7 W (weight of 24 g). Comparing these two designs head-on purely on isolated metrics we see that HP is 5.85$\times$ more throughput than AP hence should be able to process the E2E autonomy algorithm faster.

\begin{figure}[t]
\centering
  \subfloat[\# Missions.]{
	\begin{minipage}[c][0.15\columnwidth]{
	   0.3\columnwidth}
	   \centering
	   \includegraphics[width=1.0\columnwidth]{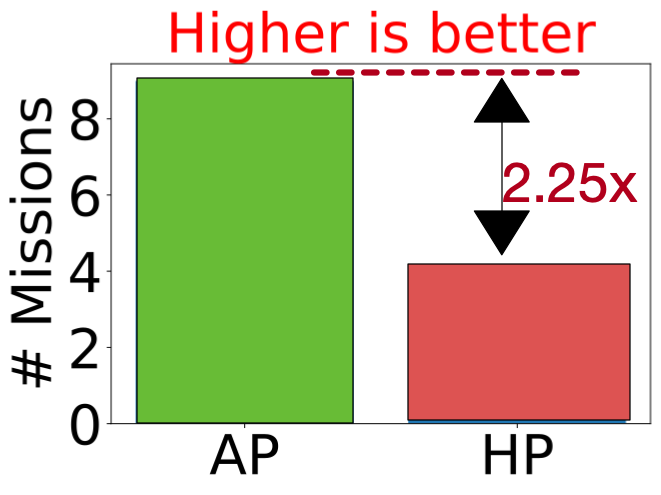}
	   \label{fig:hp-vs-ap}
	\end{minipage}}
 \centering
  \subfloat[Mapping AP and HP to the F-1 plot.]{
	\begin{minipage}[c][0.15\columnwidth]{
	   0.65\columnwidth}
	   \centering
	   \includegraphics[width=1.0\textwidth]{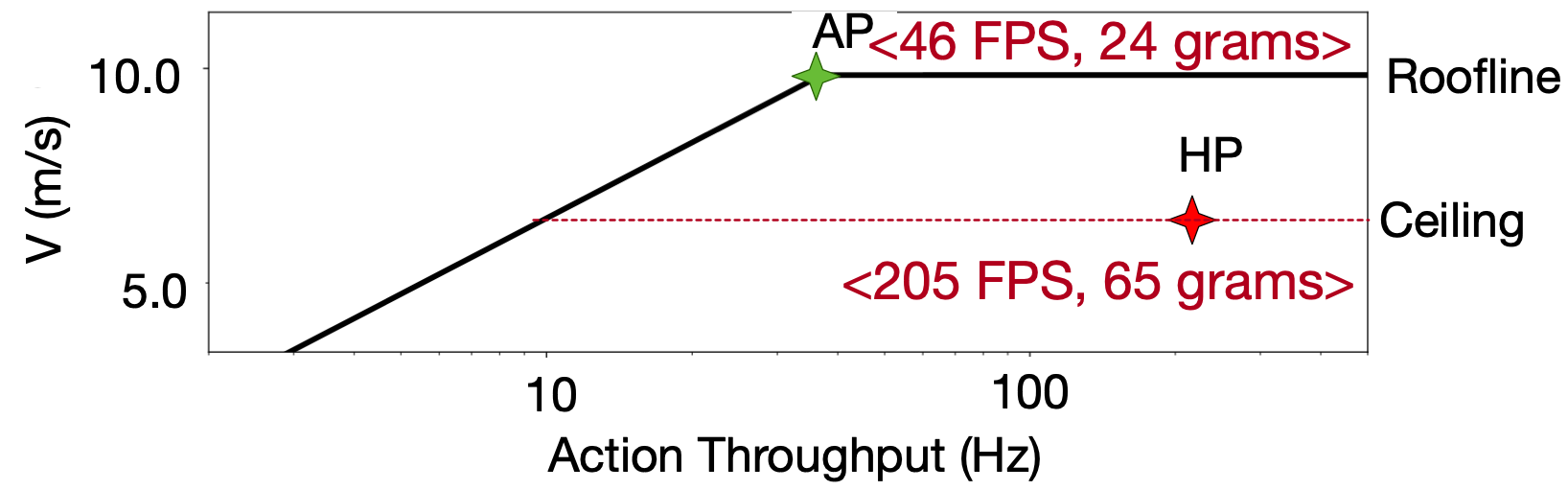}
	   \label{fig:hp-vs-ap-f1}
	\end{minipage}}
\caption{(a) Comparison of high performance (HP) and \autop design (AP) in terms of the number of missions. (b) F-1 plot for nano-UAV (Table~\ref{tab:uav-spec}) to understand the degradation when using HP over AP.}
\end{figure}

\Fig{fig:hp-vs-ap}, by contrast, shows the overall mission-level metrics of high performance design (HP) and \autop design (AP). We observe that \autop design has 2.25 $\times$ better mission-level metrics compared to high performacne design. The answer to why we see the degradation if we select HP instead of AP lies in the effect of compute weight and its impact on UAVs physics. A full-system UAV co-design (Phase 3 of \autop) accounts for these effects while selecting onboard compute design for autonomous UAVs.

\textbf{Why High-Performance Alone Falls Short?} To help explain the degradation in performance, we manually plot these design candidates on the F-1 model~\cite{f-1} for the nano-UAV (Table~\ref{tab:uav-spec}). Recall from Section~\ref{sec:cps-co-design} that F-1 is used to understand what is an optimal design for a given UAV and also shows the effects of performance of sensor-compute-control pipeline and how the weight of payload affects the UAV physics.

Since the high performance design (HP) consumes 11.7 $\times$ more power than \autop design (AP), the heatsink needed to cool HP is also larger what is required to cool AP system. This adds to the overall weight of the HP system which lowers the UAV ability to move faster (and safely). This effect is captured in F-1 plot shown in \Fig{fig:hp-vs-ap-f1}. Due to the reduction in safe velocity, it affects the mission time, mission energy, and number of missions (Eq.~\ref{eq:num-missions-expanded}) of the UAV.

Choosing a onboard compute performance based solely on compute throughput can be misleading and can even deteriorate the mission performance. Therefore, a full-system UAV co-design methodology is necessary to avoid these pitfalls.

\subsubsection{Comparison to Optimizing for Low Power Alone}
The motivation of choosing low power design comes from Eq.~\ref{eq:mission-energy} where compute power plays a role in overall energy. 
We compare the low power design (LP) with \autop design (AP) to show that \emph{low power design does not reduce the overall mission energy} and in turn has implications on number of missions the UAV can achieve on a single battery charge cycle. 

\begin{figure}[t]
\centering
  \subfloat[\# Missions.]{
	\begin{minipage}[c][0.15\columnwidth]{
	   0.3\columnwidth}
	   \centering
	   \includegraphics[width=1.0\columnwidth]{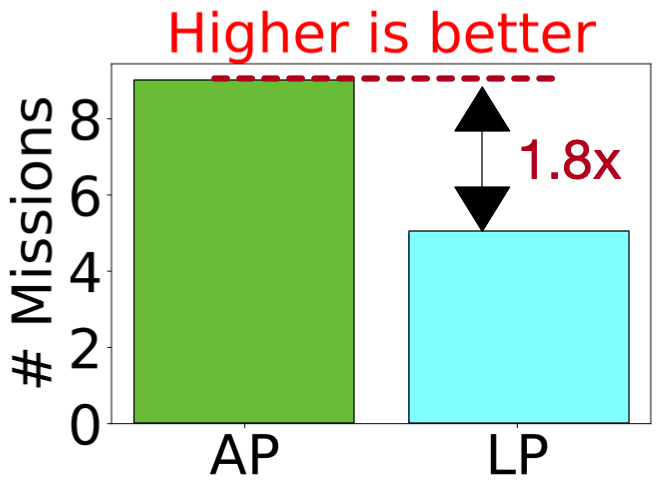}
	   \label{fig:lp-vs-ap}
	\end{minipage}}
  \subfloat[Mapping LP and HP to the F-1 plot.]{
	\begin{minipage}[c][0.15\columnwidth]{
	  0.65\columnwidth}
	   \centering
	   \includegraphics[width=1.0\textwidth]{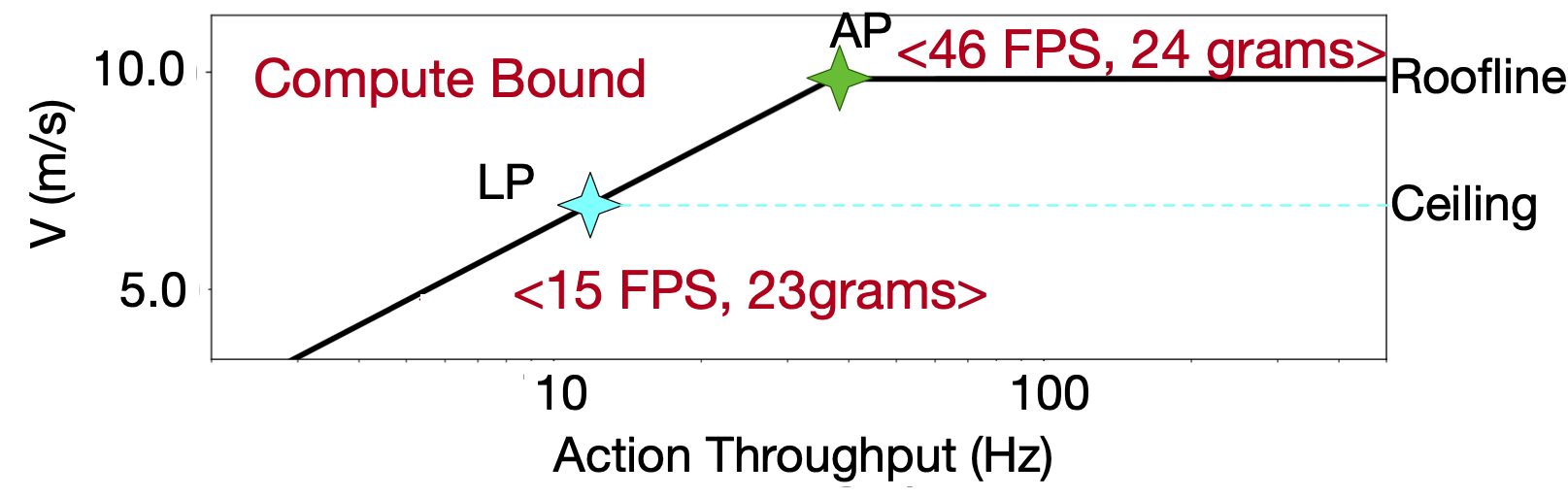}
	   \label{fig:lp-vs-ap-f1}
	\end{minipage}}
\caption{(a) Comparison of LP and AP in terms of number of missions. (b) F-1 plot for nano-UAV (Table~\ref{tab:uav-spec}) to understand the degradation when using LP over AP.}
\end{figure}

\Fig{fig:lp-vs-ap} shows the mission level metrics of LP and AP. We observe that AP achieves 1.8$\times$ more missions compared to LP.  The answer to why we see this degradation lies in the slow decision-making rate of the autonomous UAV. A full system UAV co-design (Phase 3) considers these effects while selecting onboard compute for UAV.

\textbf{Why Low Power Alone Falls Short?} To understand the degradation, we manually map low power design (LP) and \autop design (AP) designs on the F-1 model for the nano-UAV as shown in \Fig{fig:lp-vs-ap-f1}. Recall that the knee-point is the minimum action throughput required to maximize the safe velocity of the UAV. In the case of the nano-UAV, it is 46 FPS. In contrast, the LP design achieves an action throughput of which is 2.5$\times$ lower than what the nano-UAV's physics can allow. Due to lower decision-making rate for LP, the nano-UAV cannot fly safely without lowering its velocity, which affects the mission energy (Eq.~\ref{eq:mission-energy}).

Therefore, we conclude that choosing a low-power onboard compute does not necessarily lower the mission energy, since decision making rate (action throughput) also plays a role in UAV ability to fly faster, which can lower mission time and overall mission energy. Thus, compromising performance for low pow consumption can degrade mission performance and full-system UAV co-design is necessary to ensure optimal onboard compute is selected for a given UAV system.

\subsubsection{Comparison to Pareto Optimal Performance-Power}

Design methodologies co-designing HW-SW together~\cite{ReagenHAGWWB17} focus on selecting Pareto optimal designs to maximize energy efficiency. However, when designing a computing system for a domain-specific application (e.g., autonomous UAVs), these isolated methodologies do not maximize UAV performance. 

\begin{figure}[t]
\centering
  \subfloat[\# Missions.]{
	\begin{minipage}[c][0.15\columnwidth]{
	   0.25\columnwidth}
	   \centering
	   \includegraphics[width=1.0\columnwidth]{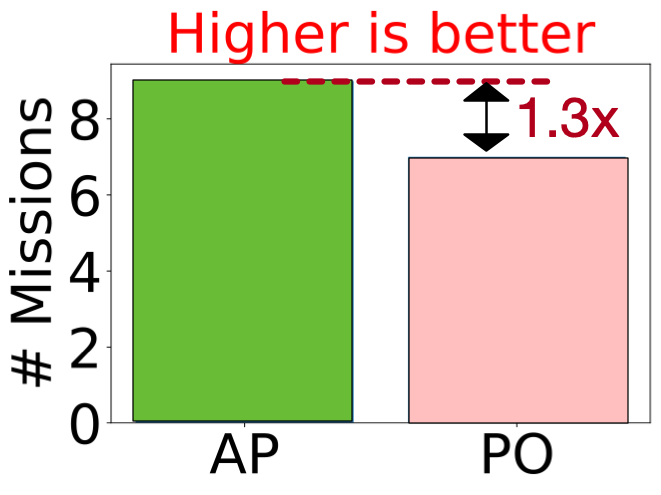}
	   \label{fig:po-vs-ap}
	\end{minipage}}
  \subfloat[Mapping LP and HP to the F-1 plot.]{
	\begin{minipage}[c][0.15\columnwidth]{
	  0.6\columnwidth}
	   \centering
	   \includegraphics[width=1.0\textwidth]{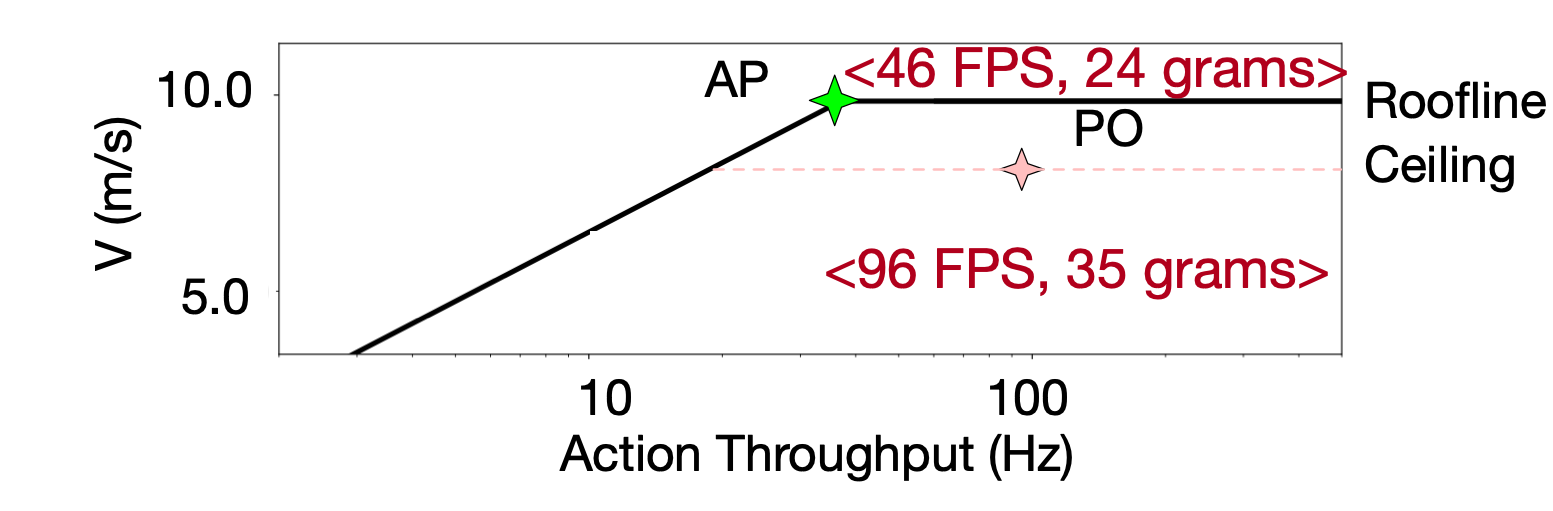}
	   \label{fig:po-vs-ap-f1}
	\end{minipage}}
\caption{(a) Comparison of PO and AP in terms the number of missions. (b) F-1 plot for nano-UAV (Table~\ref{tab:uav-spec}) to understand the degradation when using PO over AP.}
\end{figure}

To demonstrate this, we compare the Pareto optimal (PO) design with the \autop design (AP) to show that
\emph{Pareto optimal design with respect to compute performance/power may not be optimal overall} in maximizing autonomous UAV mission performance.
\Fig{fig:po-vs-ap} shows the mission level metrics of PO and AP designs. We observe that AP achieves 1.3$\times$ more number of missions compared to PO. However, despite achieving  higher energy efficiency (53 FPS/W) compared to AP (50 FPS/W), we still observe degradation in mission metrics when selecting PO. The answer to why we see these degradation lies how PO affects the autonomous UAV physics. 

\textbf{Why Pareto Optimal Performance-Power Falls Short?} The F-1 plot with both \autop design and Pareto Optimal design (PO) is shown in \Fig{fig:po-vs-ap-f1}. The knee-point design for the nano-UAV is around 46 FPS, whereas the PO achieves a throughput of 96 FPS at 1.8 W (over-provisioned by ~2$\times$). The over-provisioned PO design also consumes relatively higher power. Thus, the heatsink required to cool PO design will be higher than AP, thus increasing the compute weight for the PO design. Recall that increasing payload weight lowers the UAV's ability to move faster (\Fig{fig:f1-effects}), thus lowering the safe velocity (captured by the F-1 model ceilings), which in turn lowers the mission energy and the number of missions (Eq~\ref{eq:num-missions-expanded}).

\subsection{Effect of Sensor Performance on Compute Selection} 
\label{sec:sensor-performance}

This section shows sensor performance affects the selection of onboard compute. Design or selection of onboard compute for autonomous UAV \emph{must consider sensor performance}. Failing to account for sensor performance can cause degradation in overall autonomous mission performance.

To demonstrate this, we take AscTec Pelican (mini-UAV) with a 30 FPS and 60 FPS sensor for the autonomous navigation task in a medium obstacle density environment. For both these configurations, \autop accounts for the sensor performance and designs an onboard compute to match the sensor performance. For example, the \autop selects AP$_{30FPS}$ as the onboard compute for a mini-UAV with 30 FPS sensor. As a result, it achieves 30 FPS compute throughput while consuming 0.88 W. Likewise, AP$_{60FPS}$ is the onboard compute for mini-UAV with 60 FPS sensor. As a result, it achieves 47 FPS (knee-point for this UAV is 46 FPS) at 2.12 W. 

\begin{figure}[t]
  \subfloat[Number of missions.]{
	\begin{minipage}[c][0.25\columnwidth]{
	   0.4\columnwidth}
	   \centering
	   \includegraphics[width=0.85\textwidth]{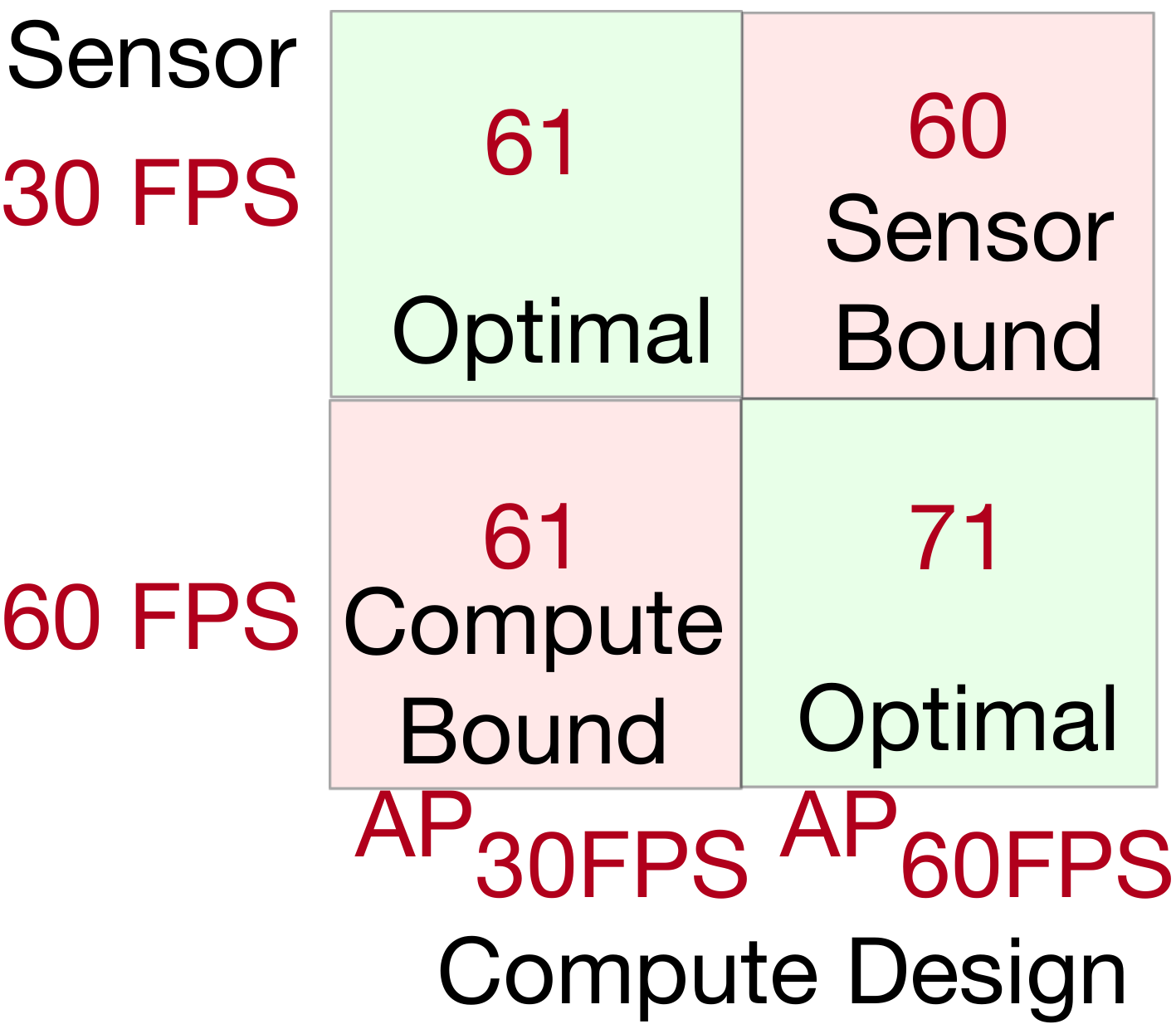}
	   \label{fig:need-cps-sensor}
	   
	\end{minipage}}
	\hspace{20pt}
  \subfloat[60 FPS sensor.]{
	\begin{minipage}[c][0.25\columnwidth]{
	   0.4\columnwidth}
	   \centering
	   \includegraphics[width=1.0\textwidth]{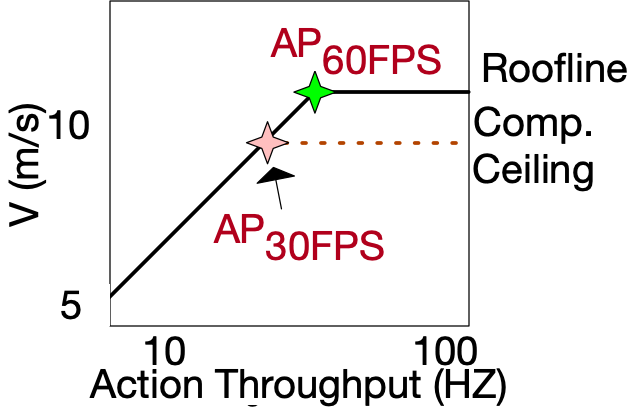}
	   \label{fig:sensor-60-f1}
	\end{minipage}}
\caption{Designing compute without considering the sensor performance degrades mission performance.}
\end{figure}

Let us also assume that for the UAV with 60 FPS sensor, we still use AP$_{30FPS}$ instead of AP$_{60FPS}$. And conversely, for 30 FPS sensors, we use AP$_{60FPS}$ design.  \Fig{fig:need-cps-sensor} shows the correlation matrix with these combinations. The diagonal of the correlation matrix corresponds to the scenarios where the onboard compute selection accounts for sensor performance. The rows correspond to the sensor framerates (30/60 FPS), whereas the column corresponds to the compute design selections (AP$_{*}$). The value inside the correlation matrix is the number of missions the UAV can achieve.

\textbf{Effect of Under-Provisioned Compute for Sensor.} In this scenario, we use AP$_{30FPS}$ design for the mini-UAV with 60 FPS sensors. Using this design point in the mini-UAV, we observe that the overall number of missions drops from 71 missions to 61 missions. To understand the reason for degradation, we map the sensor (60 FPS) and compute design (AP$_{30FPS}$) to the F-1 model~\cite{f-1} for the AscTec pelican UAV as shown in \Fig{fig:sensor-60-f1}. Since the knee-point of the UAV is around 45 FPS, the AP$_{30FPS}$ design is left of the knee-point making this design compute-bound. Even though both the UAV physics and sensor can allow the UAV to fly faster, the onboard computer's inability to make decisions faster than 30 FPS limits its ability to move faster. A slower decision-making rate has implications on UAV's safe velocity which affects the mission time, energy, and in turn, the number of missions (Eq.~\ref{eq:num-missions-expanded}).


\subsection{Effect of UAV Agility on Compute Selection}
\label{sec:uav-physics}
In this section, we demonstrate how UAV's agility impacts the onboard compute requirement to run the autonomy algorithms.
UAV's agility is typically characterized by its maximum acceleration (which depends upon the UAV's thrust-to-weight ratio)~\cite{uav-safety-2,twr-amax}. It is important to note that as the payload weight (onboard compute, heatsink etc) increases, it lowers the thrust-to-weight ratio (lowers maximum acceleration) which in turn makes the UAV less agile. Hence, there is a \emph{need to carefully account for these physical effects when designing onboard compute} for these different classes of UAVs.

To demonstrate the increase in compute requirement with UAV's agility, we take two UAV's namely DJI-spark and nano-UAV~\cite{nano-UAV}. The specification of these UAVs is in Table~\ref{tab:uav-spec}. The autonomy task for both these drones is point-to-point navigation in a medium obstacle environment. We assume that both the UAVs are equipped with 60 FPS sensors (to ensure that the sensor's performance does not bound them).

\begin{figure}[t]
  \subfloat[Number of missions.]{
	\begin{minipage}[c][0.15\columnwidth]{
	   0.7\columnwidth}
	   \centering
	   \includegraphics[width=1.0\textwidth]{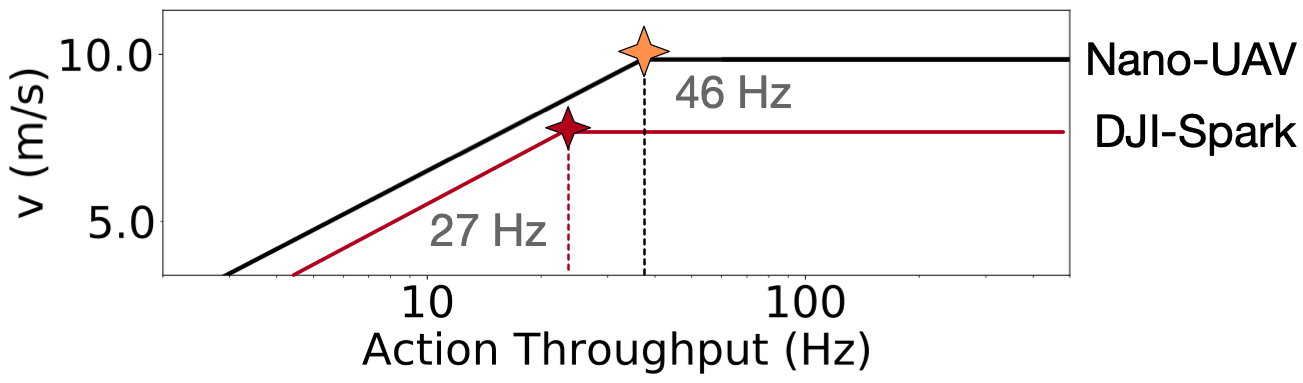}
	   \label{fig:uav-agility}
	   \hspace{10pt}
	\end{minipage}}
  \subfloat[60 FPS sensor.]{
	\begin{minipage}[c][0.15\columnwidth]{
	   0.3\columnwidth}
	   \centering
	   \includegraphics[width=1.0\textwidth]{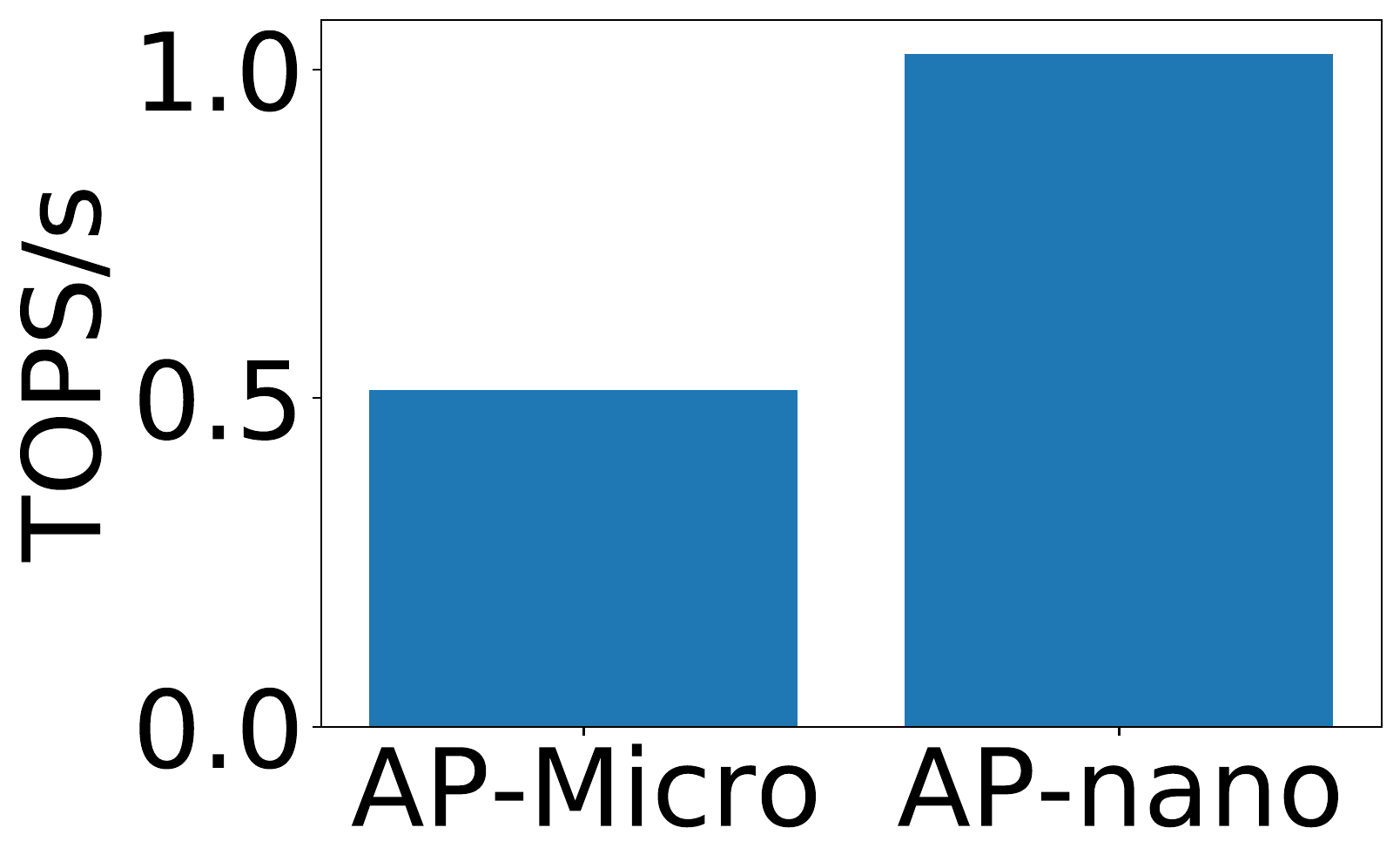}
	   \label{fig:uav-agility-ap}
	\end{minipage}}
\caption{UAV agility increases compute requirement.\vspace{-5pt}}
\end{figure}

\Fig{fig:uav-agility} shows the mapping of base-UAV configuration of both the UAVs to the F-1 model~\cite{f-1}. The optimal compute throughput required to maximize the mission performance for the DJI-spark is around 27 Hz, whereas, for the nano-UAV, it is 46 Hz. This suggests that the sensor-compute-control pipeline needs to make decisions at this rate (27 FPS for DJI-spark and 46 FPS for nano-UAV) to maximize the safe velocity. 


Since \autop uses performs full-system UAV co-design, it selects the design candidates closer to the optimal point in the F-1 model. For the nano-UAV, \autop chooses design 2$\times$ more compute throughput than the design point it chooses for DJI-spark without affecting the UAV physics (by extra payload/heatsink weight). These design points are annotated as `AP-micro' and `AP-nano' in \Fig{fig:uav-agility-ap}.

As UAVs become more agile and smaller, the role of onboard compute increases. However, minaturization also makes it susceptible to payload changes, making it challenging to design onboard compute to maximize the autonomous UAVs' performance. In these scenarios, \autop methodology is advantageous since it performs full-system UAV co-design to generate design points that maximize mission performance without impacting the UAV's physics.

\section{Specialization Cost vs. Mission Efficiency}
\label{sec:degradation}

As we show, when the UAV components change, we see the need for re-optimizing (or a new custom design). Customization for UAVs is a trade-off between its operational efficiency and design effort (cost). We quantify the trade-off between using single designs versus domain-specific hardwares.

In this case, we take the optimal point designed for a medium obstacle density scenario for a mini-UAV. We compare this design point against general-purpose design (e.g., TX2 and Intel NCS) and optimal specialized HW accelerators designed for other scenarios (e.g., Knee-point designs of Low-obs and Dense-obs but reused elsewhere). Table~\ref{tab:trade} tabulates the cost of these trade-offs. Compared to the deployment-specific hardware specialization, we observe 27\% to 67\% degradation in the number of missions possible depending upon the choice of onboard compute. Hence, if operational efficiency is important, then deployment-specific hardware specialization is key to achieving that goal. However, if cost is critical, then general-purpose designs (or reusing single designs) can save design cost with a 27\% to 67\% reduction in its number of missions (which can also increase the operational cost for grounding and recharging the drones frequently).

In summary, the trade-off between mission efficiency and the cost of computing exists, but the design complexity (cost) can be reduced significantly by \autop methodology (automating the complex cyber-physical co-design). 

\begin{table}[h!]
\resizebox{1.0\columnwidth}{!}{%
\begin{tabular}{|l|l|l|l|l|l|}
\hline
\multirow{2}{*}{\textbf{Metrics}} & \multicolumn{3}{l|}{\textbf{Single ASIC Designs}} & \multicolumn{2}{l|}{\textbf{General Purpose Designs}} \\ \cline{2-6} 
 & \textbf{\begin{tabular}[c]{@{}l@{}}Knee-Point\\ (Low Obs.)\end{tabular}} & \textbf{\begin{tabular}[c]{@{}l@{}}Knee-Point\\ (Med Obs.)\end{tabular}} & \textbf{\begin{tabular}[c]{@{}l@{}}Knee-Point\\ (Dense Obs.)\end{tabular}} & \textbf{\begin{tabular}[c]{@{}l@{}}Nvidia \\ TX2\end{tabular}} & \textbf{\begin{tabular}[c]{@{}l@{}}Intel \\ NCS\end{tabular}} \\ \hline
\textbf{\begin{tabular}[c]{@{}l@{}}\# of Mission\\ Degradation\end{tabular}} & \textit{30 \%} & \textit{0\%} & \textit{27 \%} & \textit{30 \%} & \textit{67 \%} \\ \hline
\textbf{Comments} & \textit{\begin{tabular}[c]{@{}l@{}}Compute Bound\\ lowers Vsafe\end{tabular}} & \textit{\begin{tabular}[c]{@{}l@{}}Optimal\\ Design\end{tabular}} & \textit{\begin{tabular}[c]{@{}l@{}}Weight lowers\\ the roofline\end{tabular}} & \textit{\begin{tabular}[c]{@{}l@{}}Weight lowers\\ the roofline\end{tabular}} & \textit{\begin{tabular}[c]{@{}l@{}}Compute Bound\\ lowers Vsafe\end{tabular}} \\ \hline
\end{tabular}}
\caption{\noindent Design Trade-off comparisons.\vspace{-10pt}}
\label{tab:trade}
\end{table}
\section{Related Work}
\label{sec:related}
Robot accelerator design is an area of emerging interest.
 Recent work proposed a low-power accelerator~\cite{pulp-dronet} for neural network-based control, but it is only targeted to nano-drones running DroNet~\cite{dronet}. Our work, on the other hand, provides a general methodology to generate multiple NN policies and hardware accelerator designs from a high-level specification. Navion~\cite{navion} is a specialized accelerator for improving visual-inertial-odometry in aerial robots, using the conventional sense-plan-act control paradigm. Instead, with AutoPilot we focus on end-to-end learning-based control, a promising emerging autonomy paradigm. RoboX~\cite{robox} generates an accelerator for model predictive control from a high-level DSL. Though the high-level goal of AutoPilot is similar, our work differs because RoboX does not consider the effect of the cyber-physical parameters on the computing platform, whereas AutoPilot uses the cyber-physical F-1 model to quantify design optimality. 
Hivemind~\cite{hivemind} is the first system to provide swarm-based serverless infrastructure for continuous learning on edge. In contrast, the methodology used in our work shows how to build onboard compute for a single UAV efficiently while considering a full-system view of the UAV. Therefore, we believe our methodology to maximize the efficiency of a single UAV will also scale up the overall efficiency of UAV swarm systems. Outside of the domain of aerial robots, there has been work~\cite{dan-sorin-1,dan-sorin-2,robomorphic} showing the benefits of designing custom hardware accelerators for motion planning. In particular, Robomorphic Computing~\cite{robomorphic} also provides a general methodology to synthesize custom hardware based on robot parameters (e.g., joint constraints). However, it currently focuses only on motion planning for articulated robots (e.g., arms), and motion planning is only one of many stages required to achieve autonomy. By contrast, AutoPilot provides full end-to-end autonomy for drones.
 


\section{Conclusion}
\label{sec:conclusion}

AutoPilot is a push-button solution that automates cyber-physical co-design to automatically generate an optimal autonomy algorithm (E2E Model) and its hardware accelerator from a high-level user specification.  It can be adapted to perform full-system co-design for the Sense-Plan-Act (SPA) paradigm. The only requirement for \autop is that the SPA-based algorithm and hardware templates be parameterizable. Moreover, the general methodology we have developed for AutoPilot, such as cyber-physical co-design for identifying the optimal design point, architectural fine-tuning, and selecting the optimal design points by showing how it affects the overall mission can also be adapted to other types of autonomous vehicles such as self-driving cars and ``ground'' drones.

\bibliographystyle{plain}
\bibliography{references}

\end{document}